\documentclass[10pt]{article} 
\usepackage{iclr2025_conference}
\usepackage{times}

\iclrfinalcopy

\usepackage{amsmath,amsfonts,bm}

\def\eqref#1{equation~\ref{#1}}

\def\1{\bm{1}}

\DeclareMathAlphabet{\mathsfit}{\encodingdefault}{\sfdefault}{m}{sl}
\SetMathAlphabet{\mathsfit}{bold}{\encodingdefault}{\sfdefault}{bx}{n}

\usepackage{titlesec}
\usepackage{amsbsy,amssymb,amsmath,amsfonts,amsthm} 
\allowdisplaybreaks
\usepackage{tikz-cd,tikz-3dplot,circuitikz} 
\usepackage{graphicx}
\usepackage{bbm} 
\usepackage{bm} 
\usepackage{tikz} 
\usepackage{enumitem} 
\usepackage{fancyhdr} 
\usepackage{parskip} 
\usepackage{caption} 
\usepackage{mathtools} 
\usepackage{parnotes}
\usepackage{tabularx} 
\usepackage{braket} 
\usepackage{float} 
\usepackage[dvipsnames,svgnames,x11names]{xcolor} 
\usetikzlibrary{matrix}
\usepackage{thmtools} 
\usepackage[utf8]{inputenc} 
\usepackage{url}            
\usepackage{nicefrac}       
\usepackage{microtype}      
\usepackage{stmaryrd}
\usepackage{csquotes}
\usepackage{caption} 
\usepackage{nicematrix} 
\usepackage{subcaption} 
\usepackage{algorithm2e}
\usepackage{algorithmicx}
\usepackage{array}
\usepackage{booktabs}
\newcolumntype{P}[1]{>{\raggedright\arraybackslash}p{#1}}
\usepackage{natbib} 

\usepackage{pgfplots} 
\usetikzlibrary{patterns}       
\usetikzlibrary{shadings}       
\usetikzlibrary{shapes.geometric} 
\usetikzlibrary{calc}           
\usetikzlibrary{positioning}    
\usetikzlibrary{decorations.pathreplacing} 
\usetikzlibrary{fit} 

\makeatletter
\pgfutil@ifundefined{pgf@pattern@name@_xg1qse1zm}{
  \pgfdeclarepatternformonly[\mcThickness,\mcSize]{_xg1qse1zm}
  {\pgfqpoint{0pt}{-\mcThickness}}
  {\pgfpoint{\mcSize}{\mcSize}}
  {\pgfpoint{\mcSize}{\mcSize}}
  {
    \pgfsetcolor{\tikz@pattern@color}
    \pgfsetlinewidth{\mcThickness}
    \pgfpathmoveto{\pgfqpoint{0pt}{\mcSize}}
    \pgfpathlineto{\pgfpoint{\mcSize+\mcThickness}{-\mcThickness}}
    \pgfusepath{stroke}
  }
}
\makeatother

\makeatletter
\pgfutil@ifundefined{pgf@pattern@name@_lbrsyyeax}{
  \pgfdeclarepatternformonly[\mcThickness,\mcSize]{_lbrsyyeax}
  {\pgfqpoint{0pt}{0pt}}
  {\pgfpoint{\mcSize+\mcThickness}{\mcSize+\mcThickness}}
  {\pgfpoint{\mcSize}{\mcSize}}
  {
    \pgfsetcolor{\tikz@pattern@color}
    \pgfsetlinewidth{\mcThickness}
    \pgfpathmoveto{\pgfqpoint{0pt}{0pt}}
    \pgfpathlineto{\pgfpoint{\mcSize+\mcThickness}{\mcSize+\mcThickness}}
    \pgfusepath{stroke}
  }
}
\makeatother
\usepackage{hyperref} 

\newtheorem{definition}{Definition}[section]
\newtheorem{theorem}{Theorem}[section]

\newtheorem{conjecture}{Conjecture}[section]

\newtheorem{assumption}[theorem]{Assumption}

\newtheorem{hypothesis}{Hypothesis}[section]

\newcolumntype{L}[1]{>{\raggedright\arraybackslash}p{#1}}
\newcolumntype{Y}{>{\raggedright\arraybackslash}X}

\usepackage{threeparttable}
\usepackage{ragged2e}
\usepackage[breakable,theorems,skins]{tcolorbox}
\tcbuselibrary{breakable} 
\tcbset{%
  breakable,
  enhanced jigsaw,
  parbox=false,
  nobeforeafter,
  before skip=10pt,
  after skip=10pt,
  arc=2pt,
}
\tcolorboxenvironment{theorem}{
  breakable,
  colback=blue!5!white,
  boxrule=0pt,
  boxsep=1pt,
  left=2pt,right=2pt,top=2pt,bottom=2pt,
  oversize=2pt,
  enhanced jigsaw,
}
\tcolorboxenvironment{definition}{
  colback=gray!10!white,
  boxrule=0pt,
  boxsep=1pt,
  left=2pt, right=2pt, top=2pt, bottom=2pt,
  oversize=2pt,
}
\tcolorboxenvironment{conjecture}{
  colback=red!10!white,
  boxrule=0pt,
  boxsep=1pt,
  left=2pt, right=2pt, top=2pt, bottom=2pt,
  oversize=2pt,
}

\graphicspath{{media/}}  
\usepackage{url}
\usepackage{datetime}

\title{Foundations of Artificial Intelligence Frameworks: Notion and Limits of AGI}

\author{Bui Gia Khanh\\
Department of Physics\\
Hanoi University of Science, Vietnam National University\\
Hanoi, Vietnam\\
\texttt{fujimiyaamane@outlook.com} \\
}

\begin{document}

\maketitle

\begin{abstract}
    Within the limited scope of this paper, we argue that artificial general intelligence cannot emerge from current neural network paradigms regardless of scale, nor is such an approach healthy for the field at present. Drawing on various notions, discussions, present-day developments and observations, current debates and critiques, experiments, and so on in between philosophy, including the Chinese Room Argument and Godelian argument, neuroscientific ideas, computer science, the theoretical consideration of artificial intelligence, and learning theory, we address conceptually that neural networks are architecturally insufficient for genuine understanding. They operate as static function approximators of a limited encoding framework - a `sophisticated sponge' exhibiting complex behaviours without structural richness that constitute intelligence. We critique the theoretical foundations the field relies on and created of recent times; for example, an interesting heuristic as \textit{neural scaling law} (as an example, \cite{kaplan2020scalinglawsneurallanguage}) made prominent in a wrong way of interpretation, The Universal Approximation Theorem addresses the wrong level of abstraction and, in parts, partially, the question of current architectures lacking dynamic restructuring capabilities. We propose a framework distinguishing existential facilities (computational substrate) from architectural organization (interpretive structures), and outline principles for what genuine machine intelligence would require, and furthermore, a conceptual method of structuralizing the richer framework on which the principle of neural network system takes hold. Such is then of conclusion, that the field's repeated AGI predictions fail not from insufficient compute, but from fundamental misunderstanding of what intelligence demands structurally. 
\end{abstract}

Large Language Model, or LLM (\cite{vaswani2017attention,devlin2019bert,brown2020language,zhao2023llmsurvey,zhao2024llmsurvey2,radford2019language,radford2018improving,raffel2020exploring,touvron2023llama,touvron2023llama2,chowdhery2022palm,ouyang2022training,wei2022chain,kaplan2020scaling,hoffmann2022training,bai2022constitutional}) is one of the most successful, most advanced, and most developed type of model in the current modern machine learning landscape, and of AI (Artificial Intelligence) research at large. Its success has not been lacking, and its reputation and widespread uses have been proven over time. The effect of LLM has been realized, and indeed has been changing the landscape of society in a very much difficult way. Latest model of such architecture, like \cite{openai_gpt5_2025,wang2025capabilitiesgpt5multimodalmedical,openai_inside_gpt5_2025} GPT-5, \cite{guo_deepseek_r1_2025,deepseek_hf_2025}'s DeepSeek AI, \cite{anthropic_claude3_modelcard_2024,anthropic_tracing_thoughts_2025}'s Claude AI, \cite{touvron_llama_2023}'s LLaMA, \cite{chowdhery_palm_2022,anil_palm2_2023}'s PaLM / PaLM-2, and \cite{jiang_mistral7b_2023,mistral_blog_2023}'s Mistral (Mistral-7B), have pushed this boundary further and further, and levelling up many tasks and purposes with AI system in practice, and further onward with techniques like \cite{wei2022chain}'s \textit{Chain-of-Thought prompting}, \cite{kaplan2020scaling}'s scaling law, and more (\cite{hoffmann2022training,bai2022constitutional}).

However, with such development, come great expectation, great speculation, and also great hallucination. New development in the field of AI even earlier than \cite{vaswani2017attention} paper on the Transformer neural network which fuelled the revolution of AI exposure, has gathered a group of people speculating about the further exponential growth of AI, almost to a degree of being a religious belief, about the topic of a \textit{Singularity}, where AI will become \textbf{Artificial General Intelligence} (AGI). This is reflected in popular culture phenomena, speculation, researches, interpretations of reasoning behaviours and so on, for example, in \cite{barrat_our_final_invention_2013,birch_edge_of_sentience_2024,yudkowsky_soares_anyone_builds_2025,hao_empire_of_ai_2025,bostrom_superintelligence_2014}, and more broadly on LLM in specific, \cite{mumuni_survey_llms_for_agi_2025,shang_ai_native_memory_2024,llm_cognitive_capabilities_evidence_2024,goertzel_generative_ai_vs_agi_2023,feng2024how,llms_assessment_for_singularity_2025} and \cite{cui_risk_taxonomy_mitigation_2024}. The claim is clear - we are pushing toward the age of AGI, and perhaps sooner or later, reach the state of Artificial Superintelligence (ASI) of which cited in popular cultural as the cultivation point of the Singularity - the shift of society toward a society of abundance, post-scarcity state. Most proponents point to LLM for such advancement, as it is one of the most widespread, successful and accessible form of interaction with AI systems on large scale. Push-backs against such movement, including such as \cite{friedman2024inference,lesswrong2024parrot,bender2021dangers,baan2021slodderwetenschap,towardsai2023parrots} on the "Stochastic Parrot Hypothesis", \cite{njii2024agi} questioned of LLM path to AGI, the reverend \cite{alignmentforum2025} post itself, \cite{limgen2024,limitgen2025} on generating suggestive limitation of research paper, \cite{cacm2025knockout,marcus2023elegant} critique on generative AI on world models and failure of LLM, \cite{llm2024primer} on limitation of LLM, similarly \cite{zhang2025lllms}, mathematics critique in \cite{mirzadeh2024gsm}, and more. However, the generally public, and more so of the positivity of the inner market on AI focus on the development and increment of larger models toward such goal. It is not too out of the ordinary to hear the phrase "AGI will be in $X$ days/month/year", as much as it is a social phenomena even in small or large circle. Objectively, such positivity is not without basis. Furthermore, it is rather with certain amount of irony that the research made use of AI itself, for reference taking purposes. 

Nevertheless, a critical task can be given out of such argument and thorough development of the current debate. What can then be extrapolated from the ongoing dilemma? What has to do with the architecture, the consideration about AGI that is now turned into the debate of will LLM be AGI? How is our understanding of the concept of AI, AGI, and ASI in general? And of a sense, what will provide us a pathway toward such goal? 

\tableofcontents

\section{Understanding Artificial Intelligence}

The topic of artificial intelligence has been debated since the long form of intellectual pursuit, started with, not as comprehensive starting point as possible, \cite{turing1950computing,dartmouth1955proposal}. The conversation following such cultivated into different voices of definition and opposition via problems, such as \cite{mccarthy1987general} on the definition of strong and weak AI, \cite{searle1980minds} definition of Chinese Room Argument, \cite{russell2010aima} definition of intelligence on basis of agents action, \cite{penrose1989emperor} argument against the philosophy of computational intelligence, \cite{floridi2004philosophy}'s ethics. More can be found on \cite{stanford2018ai} and further sources, but it is sufficed to say that the topic has received no less of contributions thereof to itself. However, it is reasonable to see that such argument and debate is often very shallow and hypothetical. 

Inside the term \textit{artificial intelligence}, there is the word \textit{intelligence}. A normal person will tell you that they are intelligent. But it just so happens that this notion of qualification is harder to define when one participates in the active action of finding it. So, what is it? This is the question we should take in. 

That said, this question is very much arduous in its public view, and distastes of the technical crowd. just as \cite{Oygarden2019} mentioned, for philosophers whom are assumed to be interested in such endeavour, the topic of intelligence has traditionally appeared of a less interesting concept than consciousness. Pardon their minds and view, the philosophy of mind and body seems to attract more on the front of abstract thought, than something considered to be mechanical by thinking, and attribute no more and no less to the notion of being human as it is, with the soul and body being the main question. Developments of AI have caused new philosophical interest in the concept of intelligence, though seldom appropriately decoupled from its closely related phenomenon consciousness. Even in the field of AI itself, there are the avoidances of defining intelligence, though there have been no shortage of finding one in the middle of the forest. It is said that, for someone to work in the field of artificial intelligence, it would be wise to couple oneself with a definition of AI on him/herself, rather than not. Such is to say as to fix a philosophical standpoint before working in the field, which both contributes to the enormous amount of opinionated definition, but also the rigid framework on which artificial intelligence is considered. 

As such, there exists no satisfying definition of artificial intelligence beyond the notion of artificial, of which is still dubiously believed upon. However, to fully capture the notion of AGI, we need the notion of AI on it. 

\subsection{Understanding artificial}

What separated, of the artificial and the `natural product' by definition? Below, we present the table on such terms of the concept `being artificial' taken from the most basic of the knowledge, etymology and definition available in large, of \cite{merriam,cambridge,oxford,dictionary-com,justia,etymology,bianchini,nasa,ibm,legg-hutter,goertzel} and so on so forth. Indeed, for as long as the field artificial intelligence is formed formally of 1956, in the \textit{Dartmouth workshop}, those are what we understand of the concept of artificial intelligence, as to be normalized of into triviality. In the process of making something artificial, one must then have to reproduce what they considered natural, in certain perspective --- not man-made --- into a man-made form, with certain criteria. This is discussed in \cite{simon1969_sciences,haugeland1985_veryidea,boden1987_artificial_natural_man_expanded,boden1996_philosophy_artificial_life,boden2019_from_fingers_to_digits,onyeukaziri2022_natural_artificial,boden1990_philosophy_of_ai}, in particular, about natural and artificial concept. 

\begin{table}[htb]
\centering
\begin{threeparttable}
\caption{Comparative definitions of \textit{artificial} across different domains}
\label{tab:artificial}
\begin{tabular}{@{}p{3cm}p{5cm}p{5cm}@{}}
\toprule
\textbf{Domain} & \textbf{Core Definition} & \textbf{Notes / Nuances} \\
\midrule
General dictionaries & Made by humans; imitation of nature 
& Often connotation of ``fake'' or ``not sincere''; opposed to natural. \\

Technical (AI, computing) & Human-designed systems that simulate or replicate functions of intelligence 
& Must handle unpredictable environments, learn, and adapt. Debate on narrow vs broad definitions. \\

Biology / Synthetic biology & Engineered biological systems (synthetic cells, genetic circuits) 
& Blurs the line between ``natural'' and ``artificial''; challenges classical dichotomy. \\

Classification (taxonomy) & Groupings based on superficial traits rather than evolutionary lineage 
& Used in contrast to ``natural classification.'' \\

Legal / Social & Constructs created by law, rules, or human institutions 
& Examples: ``artificial person'' (corporate law), borders, price manipulation. \\

Philosophy / Etymology & Derived from Latin \textit{artificialis}, meaning ``produced by human skill'' 
& Distinction between natural and artificial debated since antiquity; less clear in modern science. \\

\bottomrule
\end{tabular}
\end{threeparttable}
\end{table}

In all, the definition of artificial can be considered diluting, since essences and the term is usually wholly considered in different voices and perspectives. For example, one might mistakenly classify biological to being natural --- certainly a farm can be just as biological as a plain field without intervention, but not natural as it is\footnote{Under the same line, we also then have the discrete notion of \textit{biologically artificial}. A child that is born with intention of its father and mother is a natural-born child, while artificial childbirth would be a non-biological process with intervention and non-biological form to facilitate such process.}. On another matter, every object that is man-made still follows the law of physics, they obey the law of which is itself an expression of how the universe works, but such artifacts are called artificial, not natural, even though a rock still obeys such rulesets. We can then see that to equate something with the property of being artificial, requires more than just putting on its label, for the term natural itself is very hard to grasp. The science of artificial itself is now being advocated, much to \cite{simon1969_sciences} idea that a lot of the current world is now artificial in perspective. However, one can propose such dilemma by equating artificial with something entirely: \textit{process}. Simply put, we define natural as without intervention. By our scientific fact, we know that the world changes and evolves itself, develops and mutate by time, of the second law of thermodynamics. Then, we can define something being artificial as a thing created, from a process outside such evolution, of which then subjectively, means created by human. To be created by human also means to conform to the notion of \textbf{interpretation}, \textbf{encoding}, of which is then represented in certain ways or form. While philosophical, it is trivial to see that without interpretation, an object itself will remain as plain of the object itself, without purposes, without triviality. Just as a neural network without the interpretation to be a neural network is useless, and just as mathematics is useless without the interpretation of the subject and objects of consideration. Thereby, we then accept such provisional overview as the definition of artificial, from the subjective view of being created by human, hence constructible, and is created of purposes, its process and operations can be given meaning and representation. 

The definition has a drawback, though, that is, if we later on discover an alien species with the capability to construct such artificial intelligence, would we not call it artificial intelligence, because it is not made by human? We need a definition that generalize to certain notion of intended creation, for in such event if we ever discover alien lifeform, then their "AI" would not be argued against such to be not artificial because of the word's basic meaning related to human, and hence can be considered a stipulative definition in its stead. Doing such, requires the notion of \textbf{intention}, and the notion of reason. 

\begin{definition}[Artificial]
    A subject $A$ is artificial, if it is created by an object $B$, with intention and tasks $p$, constructed in a way of which the logic system of $B$ can interpret such object, the representation of $B$ can simulate and construct $A$, and $A$'s operation lies within such domain of interpretation and logic of $B$. 
\end{definition}

The notion of artificial will always be troublesome, as it is based on the ground that there exists certain subject of the current universe, $B$, that can facilitate and create another object $A$ of various means, with interventions and interferences from one's natural evolution. While not satisfactory, it will be our provisional definition on the topic of artificial. In a sense, however, it is wise to note that the definition and treatment of the term artificial is purely subjective, per subject in question. The fallacy of such provisional definition comes in the very simple thought experiment. If ants are to create its own fungus farm, would we call that fungus farm artificial, or natural? Indeed, one can attribute that to the fact that the being, observer is attributing such terms and meaning to the structure that is interpreted as fungus farm, while in fact in natural it is just a natural behaviour of the ant colony. However, how far will such interpretation stretch, and edge cases of which it can handle is unknown. Where does nature end, and where artifice begin?

Of such basis, of the artificial platform in which we as subject $B$ create, artificial intelligence theory then posit that the notion of intelligence can be represented, interpreted, and constructed using a system called \textbf{computer}. Though variations differ of which what is considered computation, how is symbolic manipulation considered, this theory is the philosophy of Computational Intelligence, or Computationalism, of which formed the basis of artificial intelligence research of date. Supporters of the theory include as earliest being \cite{putnam1988_representation,fodor1975_languageofthought,churchland1986_neurophilosophy,dennett1978_brainstorms,dennett1991_consciousness}, and of the 21st century, \cite{scheutz2002_computationalism,dodig2012_infocomputationalism,gauvrit2015_algorithmic_cognition} and \cite{copeland2018_evolution_computationalism}. This theory then posits that we can represent intelligence, of the arbitrary definition that it is understood upon, using the language of representation of computer, using numbers and computational processes as for simulation, using algorithms to simulate thinking processes, and else. Of course, there exists pushback against such notion, of which creation of artificial intelligence is not totally configurable using computers, notably \cite{searle1980_mindsbrains,searle1992_rediscovery,penrose1989_emperorsmind,penrose1994_shadows,nagel2012_mindandcosmos,muller2025_symbolgrounding,laforte1998_godel_computationalism}. To answer if the theory is false or not, and what is the implication of such theory, we have to shift to understanding the second term of the word --- intelligence. Furthermore, we also have to realize what is said as a computational structure, and as the artificial construct that we created of the computational framework, how and what will constitute the creation of artificial intelligence in such specific representation. 

\subsection{Understanding intelligence}

Inside the term \textit{artificial intelligence}, there is the word \textit{intelligence}. A normal person will tell you that they are intelligent. But it just so happens that this notion of qualification is harder to define when one participates in the active action of finding it. So, what is it? This is the question we should take in. 

We start this section with a series of historical accounts. Shane, Marcus (2007)'s paper \textit{A collection of definitions of intelligence},\cite{legg2007collectiondefinitionsintelligence}, and Masahiro's (2023) \textit{Descartes and Artificial Intelligence}\footnote{See more in \href{https://www.philosophyoflife.org/jpl2023si_book.pdf}{Journal of Philosophy of Life Vol.13, No.1 (January 2023):1-4}} might be a great place to start this, since they provide a non-trivial amount of definitions and attempts already there, which serve us more as exhibition for observant in this section and beginning. 

\begin{description}[leftmargin=2.5cm,topsep=1pt,itemsep=1pt]
    \item[R. J. Sternberg] \dots I prefer to refer to it as 'successful intelligence.' And the reason is that the emphasis is on the use of your intelligence to achieve success in your life. So I define it as your skill in achieving whatever it is you want to attain in your life within your sociocultural context — meaning that people have different goals for themselves, and for some it's to get very good grades in school and to do well on tests, and for others it might be to become a very good basketball player or actress or musician. 
    \item[D. K. Simonton] \dots certain set of cognitive capacities that enable an individual to adapt and thrive in any given environment they find themselves in, and those cognitive capacities include things like memory and retrieval, and problem-solving and so forth. There's a cluster of cognitive abilities that lead to successful adaptation to a wide range of environments.
    \item[H. Nakashima] Intelligence is the ability to process information properly in a complex environment. The criteria of properness are not predefined and hence not available beforehand. They are acquired as a result of the information processing.
    \item[P. Voss] \dots the essential, domain-independent skills necessary for acquiring a wide range of domain-specific knowledge - the ability to learn anything. Achieving this with 'artificial general intelligence' (AGI) requires a highly adaptive, general-purpose system that can autonomously acquire an extremely wide range of specific knowledge and skills and can improve its own cognitive ability through self-directed learning. 
    \item[Jensen, Huarte, Dearborn] \dots the ability to learn, the ability to understand, either principles, truths, facts, or common sense, to profit from experiences; the ability to comprehend, or the capacity to reason.
    \item[A. Anastasi] Intelligent is functionally of multiple components combined.
    \item[J. Peterson] \dots a bunch of stimuli. 
    \item[Humphreys] \dots the resultant of the process of acquiring, storing in memory, retrieving, combining, comparing, and using in new contexts information and conceptual skills.
\end{description}

Out of those definitions, there are two kinds of defining the notion of intelligence, we call it the \textbf{top-down} and the \textbf{ground-up} approach. The top-down line of thought demonstrate, most of the time conjectures, the existence of intelligence as a whole, without finding the actual shell that contains it. If intelligence is \textit{general}, then their implementation follows, but to a sufficient degree, it can be achieved everywhere. It guarantees, partially, of certain school of thoughts the generalizability of intelligence as the ground base to re-create such, which is characterized, often, by current machine learning discipline. This approach beside from guarantees such existence, also has the capability to `test' a subject of being, `intelligent'. 
\vspace{2mm}

This is done by setting up agenda and criteria, of which the current theory serves as more of a black box for the actual `machine' that contain it, but enough exhibitions fitting those criteria for intelligent. Fortunately, this also sets certain criteria for artificial intelligence to be specified so in the name. The Turing test, which posit different observable properties to be examined, and the Gödel's argument is one of such example in this line of thoughts, theorized by J. R. Lucas (1961), Penrose (1994, 1989), and Benacerraf (1967), similar to the Chinese Room Argument (Searle, 1980). Coincidentally, the notion of \textbf{computationalism} is also formed out of this approach. In a sense, it works as the following definition.

\begin{definition}[Intelligence, top-down approach]
    We say that we observe \textit{intelligence} in any given circumstances, of any arbitrary object regardless of structure, if it exhibits observable behaviours to the environment, the surrounding, the interested space such that can be clarified, and identified, to the nearest high-intelligent specimen (human), to certain degree of operational arbitrariness, of its own activities, properties, and functions. In such case, intelligence is defined per speculative reference point (human) and of criteria that fits the such point (human) model of intelligent. 
\end{definition}

The definition in the top-down sense is then entirely subjective from the human perspective. 

The \textit{ground-up} approach of defining intelligence is simply the polar opposite: Instead of defining intelligence by criteria, they create machines or models that have intelligence seems to be the emergence behaviour from those model. That is to say, they define intelligence by not defining it but constructing it. Though, this type of approach still requires the intuitive feeling of intelligence to figure out or identify such emerging signs of a growing construct, but it is more or less general, as it does not depend on certain opinion, or fixed high-level criteria to classify it. There are many ways to achieve such insight, either by examining the source of intelligence in high forms - neuroscience on human brains, or by analysing them in a representation form - as modellings, and anecdotal analogue that can be found, and so on. 

As of date, no such consensus has been found about the definition of intelligence. As we have said earlier, philosopher refrains from talking of such topic, artificial intelligence practitioners rely on certain intuitive sense and reason to interpret such intelligence definition, and some argue about such notion with terms from the discipline of AI itself. In general, it is led to believe that the overall strategy is to pick on such intuition and work with it, rather than doing much about it. And it aligns well with the philosophy of defining the notion of intelligence. 

\subsection{Artificial Intelligence}

Let us come up with an understanding of the term artificial intelligence. Make due of \cite{stanford2018ai}, it is the field devoted to building artificial animals (or at least artificial creatures that - in subtle contexts - appear to be animals) and, for many, artificial persons (or at least artificial creatures that - in suitable contexts - appear to be persons). However, such definition is fairly limited, and would not capture the essence of what practically can be artificial intelligence. Though, uncovering the current measure of which we make up artificial intelligence, we can come up with a provision definition that fits the current philosophical choice. 

For the definition of artificial intelligence, or the construct that supports it, to make sense, we need to evaluate again, from what we have seen, what is even the term. As noted by definition on the notion of \textit{artificial} in the preceding section, being \textit{artificial} mostly comes of from the consideration of evolutionary processes - of which the interaction in the physical worlds, the biological worlds, and overall, anecdotally, of anything that is non-human of its (human) own capability to morph objects into an intended state - this is what normally resided to. Then, artificial intelligence refers to a set of observations, observable qualities deemed sufficiently of all intents and purposes intelligent, by any given constructs that is created artificially so.

This breaks down to the two conceptual ways to talk about artificial intelligence.
\begin{conjecture}[Artificial intelligence]\label{conj:conjecture_AI_1}
    Artificial intelligence is the classification for any such object of constructs sufficiently reflects those qualities that fit the standard of intelligence, of which also created \textbf{artificially} of intent and purposes (as reflected in definition of artificial).
\end{conjecture}

The second conjecture then interprets that, toward the theory of computational intelligence. If such theory is indeed proved to be feasible, then we might have the following core argument: 

\begin{conjecture}\label{conj:conjecture_AI_2}
    Artificial intelligence refers to (a)\textbf{construct(s)} - of which consists of the \textbf{machine} and its \textbf{process}, for such that the machine supports the process to reflects the observed results quantified in one way or another, to be interpreted as intelligence by the construct that is standard for those terms. Those constructs however, are absent, or not, by choice, of the \textit{existential facility} - or of either a rigid static facility of such - and hence artificially made.
\end{conjecture}

Arguably, the second conjecture is far more interesting and familiar than the first one. However, the claims, of such, can be hypothesized as perhaps not so ideal generalization. The term artificial intelligence, generically, refers to the comparison between two actual constructs. If the current human - or us - are the ones evaluating certain constructs as intelligent, then it is equivalent to generalize human into a construct on its own, of sufficient analysis such that the comparison can be conducted. If so, then the definition ultimately is reinforced, as for now, to be relative and subjective. Would we be able to find generality in such structures, if of current time we rely solely on our own construct to evaluate the criteria, though the creation that we are making is inherently different?

\subsection{The Language Models (LM)}

One of the main, major example of artificial intelligence application is indeed the formulation of language, manifested in a model. Attempts has been made to trying to understand why language emerges as a proxy of information exchange, either by newer treatments, as seen of \cite{galke2022emergentcommunicationunderstandinghuman,worden2025unifiedtheorylanguage}, or per historical developments, as \cite{grimm1819deutsche,humboldt1836_language,schleicher1874_compendium,darwin1871_descent,saussure1916_course,bloomfield1933_language,hockett1960_origin,chomsky1957_syntactic,chomsky1965_aspects,goldberg1995_constructions,fillmore1988_mechanisms,fillmore1988_mechanisms,hauser2002_faculty,pinker1990_natural,deacon1997_symbolic,bickerton1990_language_and_species,kirby2001_spontaneous,christiansen_kirby2003_language,nowak2001_evolution_universal_grammar,evans2009_myth_language_universals,christiansen2008_language_shaped_by_brain,hauser2002_faculty} and more, that is both specific of the field of linguistic and broader. The ability of forming language is considered one of the many things, up to the capacity that human is capable of, hints of intelligence that human exhibits. There, it is just natural that one of the application since the early onset of AI theory, is to recreate this form of language. Some of the first major applications are, of the onset of the Cold War, the task of machine translation (MT). The first demonstration of MT, the Georgetown-IBM experiment, showed a great promise, with limited ability that was proposed to be increased even further in the future. Though, such development did not end well, and by the time the ALPAC report came out (\cite{AIWashington}) the field of MT has already been hit hard. It, and with the addition of \cite{lighthill1973_artificial} report on AI, ultimately, then officially begun the first AI winter.

Looking back as some of the failures in the theory of natural language modelling, it is perhaps surprising when looking at advancements of \textbf{Natural Language Processing (NLP)} has as the successor of such research direction in the prelude of AI research. Using analytical view upon the language, pragmatic approach to `dictionaririze' the copula of words and sentence structures (word encoding, tokenization, data analysis-like methods), simplification of words meaning, cases, categorization, probabilistic methods (for example, Latent Dirichlet Allocation - LDA, see \cite{jelodar2018latentdirichletallocationlda}), such research direction is responsible toward a huge chunk of architectures, creations of `AI language models' capable of statistically generating coherent texts and language contents, answering question in a sense, and so on, from large availability of data in text form. This is all conducted, while pay no mind into the deep theory of linguistic or the study of language itself; in a sense, a marvelous innovation, perhaps too marvelous. As because of such, some take the basis of the language model for the basis of the consciousness, intelligence emergence concept, and posit that such models, the LM or L (Large) LM, would be the centre piece of a fully realized AI, and thus, the discussion of AGI and furthermore, ASI. This is reinforced by the series of architectures that enables large-scale advancements, like \cite{rumelhart1986learning,schuster1997bidirectional,jordan1997serial,elman1990finding,lipton2015critical,graves2012supervised} Recurrent Neural Network (RNN), \cite{hochreiter1997long,gers2000learning,cheng2016long} Long Short-term Memory (LSTM), \cite{cho2014learning,chung2014empirical} Gated Recurrent Unit (GRU), sequence-to-sequence model as seen in \cite{sutskever2014sequence}, and the most foundational advanced structure of the attention-mechanism neural network --- Transformer (\cite{bahdanau2014neural,luong2015effective,vaswani2017attention}). Indeed, replicating the behaviour or coherent patterns of human language is a marvelous feat that cannot be understated. Yet, would such claim proved to be too costly, just as we have seen of criticism and empirical evidences that it is not at all omnipotent as it is pushed for? 

The main fallacy of such new approach, as will be reiterated many times, is the lack of origin, and the circumstances of Descartes's argument itself. While created such good models, it still cannot cope with logic, a wide range of logic, not simplified logic or rigid, manually designed system of logic. We are unable to determine the capacity of it to understand meaning, or any hint of such concept to exist in a language model aside from some short-lived prospect yet of no proven links to such understanding but statistical grouping. `Knowing language' does not equivalent to being intelligent, as it is always said. Furthermore, there exists a transition between language of human form, words as they are being written, to numerical encoding and manual rules on such encoding law into numerical sense, that is a problem. It brings up the question of if such models, if only the algorithm that it is, does not even understand the language itself, but is just finding the best possible answer toward the task provided. Such dilemma will have to resolve, if one is to claim language model to be the standard basis of such AI generalization. As of now, that seems to not be the case, as the cracks are closer to being revealed, and as the potential AI bubble of speculation to burst and fail in such delivery. Unquestionably, again, such development cannot be understated, and should not be forgotten or relinquished. However, pushing far beyond its weight is not a good idea of such sense either. 

\section{Understanding Artificial General Intelligence (AGI)}

We have understood the general notion of artificial intelligence, in one form or another. It is then naturally that we extend such conversation to the notion of Artificial General Intelligence, or AGI in short. In essence, what does an AGI constitute? The AGI notion relies partially on the concept of \textbf{fragmented intelligence}. This was first apparent by the apparatus of the Turing test (\cite{turing1950computing}), that suggest quantifying different human capabilities for determining a machine's ability to be `human', and as if the machine can surpass human in such range. This view is supported in a different form in \cite{gardner1983frames} book \textit{The Theory of Multiple Intelligences}, of which again posit that intelligence exists in different forms, and not a singular object of quantification. Given such, Artificial General Intelligence posits that we are able to construct, and would be able to construct, an AGI with all of such capabilities that one can consider to be human, with consciousness, with intelligent learning capabilities, with thoughts, and so on. A smaller camp, yet vocal, posits further that the structure of LLM, large language models, or \textbf{agentic AI} (\cite{sapkota2025agentic,derouiche2025agentic,schneider2025generative,wei2025agentic,raza2025trism}), will be able to achieve this goal. 

Nevertheless, the issue that plagues such notion is that the term itself is not fully understood, nor there exists any given consensus on what is the acceptable form that constitute the baseline definition of an AGI. Technically speaking, AGI can be attributed to the fact that many AI systems are constructed in fragments, of which for example, computer vision, language processing, signal processing, classification analysis, robotic spatial movements' extrapolation, and else, all of which then if can be combined into one, would inherently make a human-like form of intelligence. In between such, LLM, per its role as the language processor, is deemed to stand in between such. However, the definition in such term itself has its own fallacy. Suppose that the AGI $\mathsf{AGI}_{x}$ has multitude of ability that is inherently of its own domain and environment - computers, operating systems, software framework, etc, with full understanding and exploratory sense of such, but has no spatial movement, no sensors and the like that can attribute it of features of which human exhibits in natural sense. Would that disqualify it as an artificial intelligence, or just have to reclassify it into a different environment?

For now, we need to formalize what is AGI actually saying, in context. Or rather, a definition that is informal per its natural topic. What can be intrinsically defined to be AGI. Based on our current understanding, AGI can be defined using the basis of the fragmented intelligence theory, and by the previous statement of AI. Simply speak, of Conjecture~\ref{conj:conjecture_AI_1} and Conjecture~\ref{conj:conjecture_AI_2}, fragmented intelligence begins with considering all construct of which fits certain limited set of qualities, one or more, but distinct. Then, AGI is the plateau that there exists such construct that can fit all qualities of its quantified notion. 

\begin{definition}[Artificial general intelligence]
    Based on Conjectures~\ref{conj:conjecture_AI_1} and~\ref{conj:conjecture_AI_2} and the fragmented intelligence scheme, define a quality basis $\mathsf{AI}_{n}=\{A_{1},\dots,A_{n}\}$ of arbitrary given qualities specified of an artificial intelligence construct. Then, a construct $\mathsf{AGI}\in\mathsf{AI}_{n}$ is called an artificial general intelligence, if $\mathsf{AGI}$ can be represented as 
    \begin{equation}
        \mathsf{AGI} = \alpha_{1}A_{1} + \alpha_{2}A_{2} + \dots + \alpha_{n}A_{n}, \quad \alpha_{i} > 0, i = 1,\dots, n
    \end{equation}
    That is, $\mathsf{AGI}$ expresses every given qualities of the quality basis, to a given evaluation degree $\alpha_{i}$ associated with each quality. Thus, we say $\mathsf{AGI}$ spans the entire quality space. 
\end{definition}

Here, the definition is arbitrary on purpose, as for many evaluation metrics, testing setup, or different quality that is regarded of AGI, for example, the updated version of the Turing test versus the traditional one, we resolve to such definition by the action of generality. Of course, this begs the question if certain quality basis $\mathsf{AI}_{n}^{i}$ is better than others, and the answer is yes, but subjectively --- since the qualities themselves usually cannot be quantified exactly. Nevertheless, we posit that such construct can exist, from the construct's structural system itself, that when mapped of its operation onto the arbitrary basis of choice, yields such resultant observation. There then exists the fundamental problem of such definition --- for such artificial intelligence construct to exist, within the fragmented theory of intelligence, then there must then exist internal connections between its components, and the expression oward separated qualities' evaluations. Because such processing connector also retains its own quality specification, in such framework, aside from being a hidden, internal component of the model (of which then handle such connection arbitrarily and cannot be evaluated simply), it must then be of certain construct that is present within the basis. Such is usually reserved for large language model, as it is on the basis of natural language qualities, and is naturally the candidate in such regard as the main focus of core component for determining the feasibility of an AGI. Such framework is then called, the \textbf{agentic AI} framework. By certain arbitrary consideration, we can say current AGI, of certain qualities such as the Turing test, is already an AGI, within the above definition. Nevertheless, by such basis, we can set the bar so low that such AGI is unbearably naive, or set the bar too high such that the qualities in consideration is considered an infeasible event by `superhuman standard'. Or by a list of all details, measurable set of qualities with overlapping, yet does not and could not usually take into account of the internal structures of the model and the inherent natural abstraction, or emergence of whatever definition being the arbitrary emergence, is for such given construct. In a sense, we already have AGI. The problem is how effective it is. It is perhaps natural that we re-question the anecdotal concept of LLM, and of the basis, and of the entirety of connector framework, and extrapolate from such --- what can be different from LLM and such current understanding, and what is inherently wrong and right with it?

This view is debated fiercely, as we have introduced it before, and also because of its intrinsic nature of the field itself. For the example, the arbitrary of such consideration is a problem as there would imply the problem of inconsistency in between criteria, qualities, and so on, of which has been surprisingly apparent as many times standards and qualities have been modified and pushed forward in the case of determining "What is then even A(G)I?". However, there are also voices aside from such camp, telling us that current structures of artificial intelligence understanding, machine learning anecdotal knowledge, are enough to construct such AGI. Some further argue that the current framework is complete, of which provide exponential growth of such toward AGI in such matter. What is the correct answer to such question? Of this paper, the current stance is no. We then have to present logics and evidences to back such answer up. 

\subsection{The fallacy of defining intelligence}

We mentioned the notion of the criterion of intelligence. However, what should we define it? How should we know to even evaluate it, is a very hard question even that we did not (or unable to) fully realize yet, then what we want to do with it? This question is where a lot of things in the artificial intelligence research was based upon. For example, the (Total) Turing Test in which outlines possible outlook for intelligence, for capabilities that then defines the fields in which we are having nowadays, for example, computer vision for the capability of visual perception, natural language processing (NLP) for the capacity of language, and more. We also have various conceptual criterions in which people have been suggesting about the model of the intelligent being, for example, various set of criterions that outlines and includes even consciousness, some suggest behavioural conditions, some goes for the exhibition of \textit{chain of thoughts}, and some even goes further than that, which is perhaps irrelevant aside from mentioned for example. Overall, it is perhaps a mess. 

We still do not know what to come of criteria, or rather, in the quest of producing intelligence, we base ourselves onto it too much. As a species capable of intelligence and more sophisticated notion, we have the basis, and the advantage of being able to examine ourselves. By that, eventually, as the highest example of intelligent being, we use ourselves as standard, for examples, of psychology, neurological behaviourism, neuroscience, applied onto the quest of going for artificial intelligence. Hence, there exists the total Turing test, and there exists the conflicts between various definitions and criterion of artificial intelligence. A mistake perhaps has been made, doubtfully so that one did not realize of such. While it is said that AI researcher has been working on, or at least researching on the general notion of artificial intelligence principles, it is, in fact, not so much of a principle, as we do not realize, yet, that what we are doing is still the act of mimicking ourselves - creating a plane by replicating a bird. By phenomenologically absorb and construct architectures, models on the higher-level surface of what artificial intelligence constitute, the deeper construct is still non-existent. By copying the apparent capabilities of human and related intelligence being, biological rather than not, the core of which those behaviours occur, and facilitate the organs and observations made is perhaps, manifested, simply does not exist. Ironically, while being too strict, wrongfully abhorrent to the fallacy of themselves, and too resistant to changes, symbolic approach got one of the right thing. If there exists intelligence, then it must be \textit{universal by virtue}. That is, you cannot argue that alien from another universe is not intelligent, because they do not satisfy one of the criteria of the Turing test, just because such notion does not exist in such universe. 

Historically, it does not prevent people in the field of artificial intelligence to search for intelligence in their bold claims, of which their arguments make clear that they are phenomenologically mimicking the attributes of which is of human, without the substances underneath: 

\blockquote[Pollock, 1995]{Once (my intelligent system) OSCAR is fully functional, the argument from analogy will lead us inexorably to attribute thoughts and feelings to OSCAR with precisely the same credentials with which we attribute them to human beings. Philosophical arguments to the contrary will be passé.}

From such view, I objectively don't think we should, or we could define artificial intelligence, at least of this particular stage that we are in. Philosophically, being an armchair philosopher would not help in pursuing such notion, yet again because we are arguing on the basis of our own existence, and not the subject's matter viewpoint. There are problems related to it, also, of such that the mind and consciousness is arguably debatable in every given sense, of which no one seems to agree on the mundane notion that intelligence and consciousness come from chemical and the weird 'quantum effect' that would be then believed to be. And, truth to be told, we are not even endorsing such direction. In actuality, we don't even know what is intelligent, and also don't even know what can be of artificially made rather than matching mathematics. 

On the flip side, computationally and neuroscientifically, the lack of formal treatment and overall encompassing knowledge conjunctions plague the construction and foremost attempt to do anything, simply because too many things have been said yet none can unify them together. Such is also to say different directions and different methodologies being conducted, yet they are so distinctively separated to be unable to conform one to another, despite them taking on the same object. Furthermore, there are a lot of assumptions given in computational theory, and the overall application thereof. As for anything, assumptions can be broken, and reinforced, for whatever it is being inconsistent as a virtue. 

It is wise to remember that, for now with neuroscience being not advanced enough and in a perhaps different direction from what can be seen, while certainly for empirical science we can utilize neuroscience's knowledge, we should not take in the philosophical arguments and 'idea', including computational theory of mind. For empirical neuroscience, it is also not the fully-encompassing field that observe the brain from every angle, and observe consciousness of everything if ever, at least of the present. And, for the \textit{philosophical} and idealistic view, only one thing can be said about such being "the lines on the map is made up". 

\subsection{The statistical model critique}

AI, as of current, have shifted its structural formations and logical acumen back to mathematics. That is, right now, artificial intelligence looks like nothing but the thing it is originated from, but rather statistical models on data. This view is iterated in several literatures, of which we list of \cite{penrose1997artificial,peyres2025mathematics} and \cite{kutyniok2022mathematics}. While not downplaying the role of mathematics in its application and advancements of the field, and successful creations such as many models have been created, one question remains - is it actually artificial intelligence, or just a statistical model trained and probabilistically interpreted to mimic certain aspect or tasks of which is considered intelligent? This is also the stance that \cite{searle1980_mindsbrains} argued against, of which produced the long-standing Chinese Room Argument. We can summarize the argument simply as followed. 

\blockquote[Searl, 1980s]{CRA is based on a thought-experiment in which Searle himself stars. He is inside a room; outside the room are native Chinese speakers who don't know that Searle is inside it. Searle-in-the-box, like Searle-in-real-life, doesn't know any Chinese, but is fluent in English. The Chinese speakers send cards into the room through a slot; on these cards are written questions in Chinese. The box, courtesy of Searle's secret work therein, returns cards to the native Chinese speakers as output. Searle's output is produced by consulting a rulebook: this book is a lookup table that tells him what Chinese to produce based on what is sent in. To Searle, the Chinese is all just a bunch of --- to use Searle's language --- squiggle-squoggles.}

It is notable to point out that Searle's argument against the current advancement of AI using CRA is particular still effective even in the modern current landscape of artificial intelligence. In fact, this voice resonates with a large pool of people, whether because of the fear of losing identity, or from analytical assessment of current models. Large Language Models and advanced models often fail miserably when changing context or changing setting, losing information or hallucinating and so on for a large spectrum of situations, for imperfections that lies outside its intended encoding. Coincidentally, this also fit the old-age argument that Descartes made on the notion of machine and human, or simply the Descartes's argument (\cite{Descartes1950-DESDOM}). Descartes's argument start with the justification of the apparent reactive behaviours observed by human themselves, who at the time, was largely considered to be the only species capable of advanced rational thoughts and processes. 

\begin{displayquote}
    (I)f someone touched it (= the machine) in a particular place, it would ask what one wishes to say to it, or if it were touched somewhere else, it would cry out that it was being hurt, and so on. But it could not arrange words in different ways to reply to the meaning of everything that is said in its presence, as even the most unintelligent human beings can do. [Descartes, 1700]
\end{displayquote}

Here, Descartes argues that in order for human-like robots to acquire intelligence, they have to gain a universal capability to accurately react to any unknown situation that may happen in the environment. However, what machines can do is no more than to respond to a single situation one-on-one via a specific organ, hence, they cannot be considered to have a universal capability that even unintelligent human beings can enjoy. 

Continuing, Descartes argues that those machines do not act on their knowledge, but the disposition of organs.

\begin{displayquote}
    For whereas reason is a universal instrument that can be used in all kinds of situations, these organs need a specific disposition for every particular action. It follows that it is morally impossible for a machine to have enough different dispositions to make it act in every human situation in the same way as our reason makes us act.
\end{displayquote}

The argument is quite clear. Human is universal of the environment. Whereas machine is no more than a combination of abilities that are applicable only to certain situation that the creator could imagine when they built the automated machine. This simply posit, and be relevant of our observations on the current machine learning models and large language models. While they are large and as a result, of great range of capability, they are inevitably hard-coded with what the designer wishes them to do. They are not intelligent, in a sense, so to speak of their capability as to be artificial intelligence of its true `general form'. Hence, of this debate, we can simply say, to the disdain of the empiricists themselves, that artificial intelligence of the current structure simply is not distinguishable from the statistical model view. 

The same plagues the new-found sensation of agentic AI, of which is believed to obtain general intelligence by cutting and wrapping different small specialized-AI structures together, for example, connecting an LLM to a computer vision system or image recognition for scanning documents. However, by default, such connection is inherently shallow, as there exists only a \textit{black-box connection} of input-output, process resultant in between those components together, which makes it similar to CRA and Descartes's argument in question. 

\subsection{The intelligence model}

Along the same line as the statistical model critique, we now change to the perspective of the current architecture of choice. Assuming the current knowledge of artificial intelligence and thereof, within respect to machine learning and other development fields, can we say that we understand, or at least can construct artificial intelligence, and hence the true form called AGI? The answer is both yes and no. 

Current understanding of AI, \cite{10.5555/1671238}, as specified, focus on the self-reflection of the human researchers on themselves. Such reflection are often surface-level, for example, behaviours of which intelligent choice might occur rather than not, situation of which there exists patterns in which the mind choose to operate, and so on. However, this in particular, face heads-on with a problem that even now cannot be explained or go through - the domain problem, or \textbf{the frame problem}. The frame problem is the problem that an AI cannot autonomously distinguish important factors from unimportant ones when it tries to cope with something in a certain situation. The problem arises, for example, when we let AI robots operate in the real world. This problem was proposed by John McCarthy and Patrick J. Hayes in 1969, of which is considered a philosophical problem that cannot be merely reduced to a technical problem. Historically, the problem is narrowly defined for the field of \textit{logic-based artificial intelligence}. But it was taken up in an embellished and modified form by philosophers of mind, and given a wider interpretation, and hence, is since then applicable to almost all formal system that wishes to call themselves artificial intelligence. We will not cover all of it here, at least for now. For authoritative literature, it is recommended to refer to the Stanford's article on the frame problem \cite{sep-frame-problem}, and other literatures \cite{FrameGryzJarek,SeagerFrameAxiological,Briggs2014MachineE}. However, it is indeed a dilemma of which both symbolic Ai and the kind of statistical, variable AI of current form cannot proceed. Even with the structure of neural network and modern deep learning, symbolic encoding, expert systems, current AI structures are what called \textit{interpolator} in the purest form, and not extrapolator. This problem is inherently similar to how we would interpret the problem of \textit{out-of-bound} cases are, as seen in \cite{bahng2022learning,xu2020how,traber2020relex,huellermeier2021aleatoric,amini2021quantifying,sensoy2024epistemic,theresa2022information} of a wide range of such notion. Additionally, there are many problems that would not have answer and thereof, for example, the problem of \textit{hallucination} apparent in typical LLM settings, symbolic grounding problem in \cite{Harnad90SymbolGrounding}, and so on. The problem is not with identifying the problem and observing it --- the problem lies mostly in the form of \textit{inexplainable phenomena}. To do such, we have to inquire further into what is being done in the current theory landscape.

\subsubsection{Artificial Intelligence Theory}

The current artificial intelligence theory can be branched into several aspects, as seen in \cite{Jackson2019,GoertzelPennachin2006,Chowdhary2020,10.5555/1671238}, and more. Those considerations of the field are what we can call as both phenomenological inspired systems, and general modelling theory. We can list a few of them, for example, \cite{shortliffe1976computer}'s MYCIN structures, \cite{newell1976computer} works on symbol systems, \cite{collins1969retrieval} semantic memory, \cite{davis1977production} knowledge-base representation, and \cite{Lindsay1993DENDRALAC} DENDRAL expert system for working functionals of chemical compounds. There core driven such developments are the artificial intelligence idea focus on the reasoning process of the brain, using either pure logic (Logicism), Expert Systems, Non-monotonic Logic, Planning Systems, Argumentation, Semantic Network/DL, and Modal/Temporal Logic; more can be found in \cite{math13111707}. 

Theory like knowledge representation, agent structures, algorithmic searches, decision theory, rule-based or relational network-based reasoning theories, first-order logic, specific domain of thoughts of computationalism, symbolic AI, or neurosymbolic unifications, while proved to be effective in a sense, those theories do not resolve the origin problem for those behaviours, and furthermore is restricted in the same restrictions that we stated as above, of AI models in generality. Logic in the construct of symbolic approach deliberately catches surface logic, representations, knowledges, properties and attributes, and assigning strict logical conclusion together for the logic itself. This falls into the range of Descartes's argument, in which one's machine deliberately can only follow and operate of its designer's configuration, nothing more and nothing less. The restriction of a decision tree can also be founded to be limited in such case, for there are limitless consideration and fluctuated information in a given setting. Furthermore, a decision network on itself does not have any meaning. For example, the programming language \texttt{PROLOG}, which is prominent of such framework design (for reference, see \cite{clocksin2003programming,colmerauer1972systeme,colmerauer1993birth} for details on the standard of \texttt{PROLOG}), different facts are encoded manually by \textit{atom}, a unit of logical fact in the database, and interpreting symbolic tasks being the goal, of which \texttt{PROLOG} then traverse the symbolic graph encoded to find the solution, of which also return, in the same sense, an encoded answer. 
\begin{figure}[htb]
    \centering
    \begin{subfigure}{0.4\textwidth}
        \centering
        \includegraphics[width=0.8\textwidth]{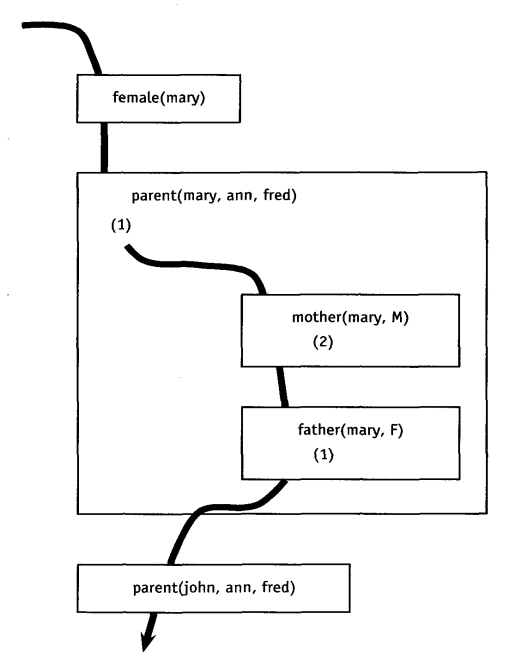}
        \caption{Resultant process.}
        \label{fig:subfig1}
    \end{subfigure}
    \hspace{1pt}
    \begin{subfigure}{0.5\textwidth}
        \centering
        \includegraphics[width=1.1\textwidth]{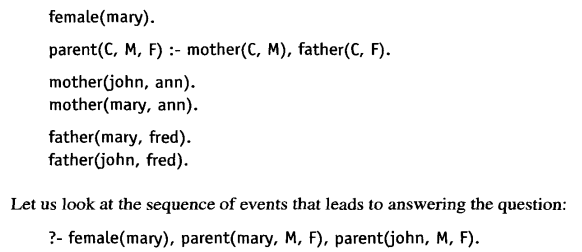}
        \caption{Program.}
        \label{fig:subfig2}
    \end{subfigure}
    \caption{A typical example of a program and its logical process in \texttt{PROLOG}. The example requires determination of a parent-child relationship, with encoded knowledge graph, and traverse of the predicate logic. (a). Canonical example of a \texttt{PROLOG} predicate decision network, in an example of Chapter 2, section 6 in \cite{clocksin2003programming}; (b) The program responsible for the respective sequential predicate logic, of which answers for the question accordingly.}
    \label{fig:overall}
\end{figure}

The program, while succeed, only works in its own environment. It is simply a program with encoded ruleset, of configured system of objects, and would rather be classified as an algorithmic program than artificial intelligence. Such parent-child relationship only makes sense in the eye of the designer, yet questionably non-apparent of the program itself. It is similar to the CRA, in which symbol manipulation and fact encoding only get it thus far, without any sufficient notion of understanding, for the arbitrary definition of such term is defined. Furthermore, applications using such system is severely limited at scale, for manual addition of facts, expert systems (from expert knowledge and so on), of which add into the rigid mimicry, and many more design initiatives. 

When the focus switch to another constructive structure of connectionism, of which is utilized in statistical and data-driven frameworks, models like probabilistic logic, statistical relational learning, Markov/Bayesian process network and logic, Causal Inferences, Deep Neural Networks (DNN), and so on, they also struct the same disadvantage as discussed of the majority of symbolic-based program, sometimes worse. In terms of mathematical modelling, \cite{VeltenetalMathematicalModelling}, such models in statistical framework and on are interpretations of a \textit{phenomenological model}, or \textit{black-box model}, in which data and observations are encoded as statistical points for interpolation, either soft (for $\epsilon>0$, the disparity between the wrong answer and the correct one is not required to be reduced to zero) or hard (the condition explicitly said to reduce to absolute 0, or point-fit), without any knowledge of the underlying mechanical details. Hence, they are usually nothing but statistical fit of particular problems, for example, classifications, regressions, predictions, and so on of numerical data encoding. Sometimes, the encoding range and model structure is extended so hard, that the model seems too real, as seen of LLM, yet is not real, by the observable statistical-inherent errors and limitations. A limited observation can be said, as in \cite{song2025languagemodelsfailintrospect}. 

The flaw of the current artificial intelligence theory can simply be attributed to looking at the wrong way, of which while proved insightful, is misleadingly taken as the way forward of making an intelligent construct. Approaches and constructions listed above relies on the basic idea of mathematical modelling, the modelling theory, yet such theory is underdeveloped, hence the true nature of the modelling scheme, of the generality on the theory of the representation language of which represents and is used for descriptive construction and analysis of models, often not realized. Such is also seen of many theories that is of apparent importance to artificial intelligence advancement, yet is rather forgotten in pursuit of the state-of-the-art phenomenological models. Different conceptual understanding, while making sense and is rigorous of logic and foundation, often found itself in trouble of interpreting the strong argument about origin of such system, once ask of the designer to make the model learn of the concept by itself. As of now, only in the sense of mathematical reduction, data points reduction that the notion of learning is then realized, and yet such is insufficient to be called intelligent, but intelligent algorithm. There exists no unifying or general theory of which is concerned of the topic of encoding, representation, architectural design and implications, operation theory, percolation theory and systematic emergence and thereof, which makes any inquiry into the topic harder than it should be. Current theories also use their limited systems to make bold claims, for example of manifesting consciousness, emergent behaviours, models as `living' while such criteria of living is not fully realized\footnote{Each pass to the model is a statistical or logical process, of which is almost algorithmic-based. This is arguably living, since they are active only by designed activation, patterns of activation that is concrete, and of operations that would not be made apparent of its meaning, for meaning is not strictly defined.}, making it inconsistent, and created a skew trend toward practical, empirical construction but no theoretical understanding. Again, it does not prevent us from making use of it, for applications on such first ever dynamic system modelling is enabled, opening a wide variety of different structural encoding and thereof. But even then, the theory of such system itself is insufficient, of limited depth, or coherence that is typically seen of a matured field. 

Other than such, treatments of empiricism on such artificial intelligence theory, either by reconstruction of the brain for neurophysiology and neuroscience, empirical constructions and continual improves, such as current model development of AI systems, remains deeply shrouded in unanswered questions, inexplainable behaviours, constrained results or returns, high computational cost of inefficiency, scaling issues, and many more. Of such, it seems questionable to continue pushing toward that end, as the theory has exhausted a fair lot of its development, without gaining much insight or exploratory evidences into the deeper questions about AI, to ever reach the node of AGI. On the philosophical side, many has argued against the notion of computationalism, whether computers can actually be used as the foundation to create intelligence form, and understandably yet unsurprisingly the current theoretical treatment cannot have an answer toward such argument, other than keep working on it until a roadblock is hit. 

\subsubsection{The learning theory}

In all of artificial intelligence theory foundation, one if not the most important aspect, and hence field of research that now dominated the field, is the theory of learning. Inherently, from the onset, the particular intelligent behaviour that one can instantly attribute to a construct to be called intelligent, is the ability for it to learn. The notion of learning is difficult to define, hence current theory seek to determine the mechanical equivalent in mathematical form instead, of which is expressed using the current \textbf{machine learning theory}, \cite{10.1145/1968.1972,wolpert1996lack,shalev-shwartz2010learnability}. In between the learning theory, there exists the fundamental main side, one of which is concerned with pure logical encoding, or rather learning machines of symbolic logic with applications in a wide range of different fields, as seen in \cite{1056797NewellSimon,newell1959gps,winograd1972understanding,michalski1969aq}, \cite{minsky1961steps,minsky1968semantic}, and \cite{quinlan1986induction}. On the other side, is the looser end of the learning theory, which seeks phenomenological learning through enough inferences onto data, or rather, information-based learning, attributing learning to pattern recognitions on \textbf{observables} of specific problem set, as seen of \cite{solomonoff1964formal1,solomonoff1964formal2,gold1967language,newell1958elements,cover1965geometrical}, and its formal treatment in theory by \cite{vapnik1968uniform,Vapnik1999-VAPTNO,vapnik1971uniform,valiant1984learnable,angluin1988queries,computationallimits1989,board1992necessity,STL_Hajek_Maxim_2021,10.5555/2371238,10.5555/2621980}. 

Learning theory is regarded strongly as one of the most important part of a model in the modern modelling landscape. In essence, it is considered detrimental to be able of lifting up the problem of which previous versions and generations of symbolic and rule-based AI, much because of their rigidity, stumbled upon. Hence, the learning theory often makes use of the notion of statistical phenomenological approximation, and different methods to construct models and expression that take into accounts information, deviations, noisy settings, and thus can be applied more generally and of specific setting accurate of observations. 

These two are fundamentally different approach, yet of the same goal of learning, or adaptation of reasoning and \textit{unseen situation} of which then previous knowledge can be applied in certain way, or by extrapolate new information using a copula of approach, either by observation or by deductions. Despite their success, symbolic learning is plagued with its own originality --- it cannot be scaled effectively, it is too rigid for such understanding to be made, and it is inefficient in handling truth or logic outside its rigid domain. Statistical learning, or rather learning by data in its pure form, relies on the model of \textit{black-box} modelling approach, usually not totally by can be seen of as the \textit{grey-box} modelling, which has its own problem about interpreting the concept wrongly, surface-level phenomenological approach which leads to insufficiency of mechanical details that underlies the concept that it learns, the reliance on data and large scale observation sets, while being scalable remains deeply mystery, of which cannot be explained (so-called inexplainable AI/DL), unanswerable questions regarding the different phenomena and computational process, results and performances, additionally structural interpretation, and so on. Furthermore, the theoretical treatment itself is fragmented, with many theories competing for interpretation and explanation in the wide spectrum, and with many different non-unified outlook of the same system dynamic, for example, \cite{mcallester1999some,alquier2008pac,haussler1996rigorous,jeon2025informationtheoreticfoundationsmachinelearning,abreu2025topologicalmachinelearningunreduced}, and so on. For the large majority of current theoretical treatment of machine learning, it is stuck interpreting a fixed domain of result and setting, reliance on interpretation of the statistical parameterized model $\theta$, and using learning strictly in the domain of optimization theory only. Newer frontier in architectural researches, such as modern deep learning \cite{marcus2018deep,goodfellow2016deep,10.5555/2721661}, proves tremendously difficult to fully understanding using such theory, as for many of unanswerable questions were born from such architecture alone. The ergonomic and logistic of a unified theory of learning, either both symbolic (experts' voice of unifying both phenomenological learning with symbolic knowledge) has become apparently so large in the eye of practitioners that some inherently try to create new theory of interpreting the learning framework, or push further in some directions regarding such, of which then again, mimic the 1970s-1980s age of artificial intelligence discovery, where many interpretation, exotic AI systems, and prototypes exist far and large. 

It is undoubtedly successful of advancements and technological leap that the current theories landscape has provided with regard to artificial intelligence knowledge, for example, of figuring out a system better in some regard, human counterparts. Yet, it is clear to see that such theory is not capable of handling various degree of new problems, of new consideration, and of new standards of which has been hidden from the main argument of AI development, the question of interpretability, the question of originality, many questions of behavioural concepts, such as the question on "common sense learning" (while it is foolish to study this at this stage). It is clear that we either need a unification of all such learning concept, and a new development of a more powerful learning framework, or to jump the bandwagon entirely; or something else. 

The main critique also lies in the heart of the current implementation of the machine and its learning process as of date. Indeed, one can see a learning process as not the machine itself, but as two separated entities: the pair $(\mathcal{A},\mathcal{M})$ of the algorithm $\mathcal{A}$ for the process of learning, and the model $\mathcal{M}$ itself. While being considered such, it is in fact concerning, at least in certain part of the dialogue between the camp of learning theorist, and others, is the claim of which the machine or model itself is already learning, while in fact it is not. 

\begin{figure}[htb]
    \centering
    \includegraphics[width=0.7\textwidth]{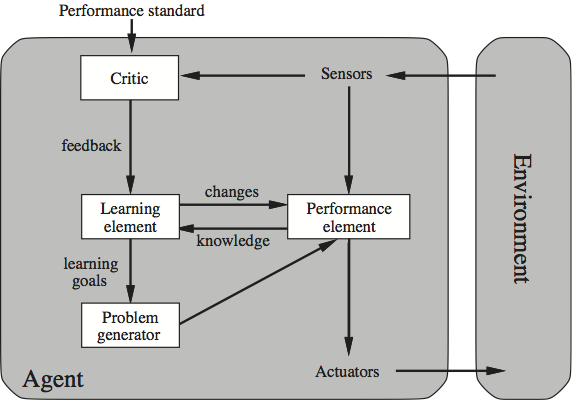}
    \caption{The conceptual framework of a learning agent (\cite{stanford2018ai}). The agent itself in such framework then indeed, internally `understand' the metric notion of mistake, knowing and then fixing, typical process of which a learning system can be considered.}
\end{figure}

As such, for those argued that machine can already learn and think, it is instead a very sophisticated pattern matching and statistical inferencing process. In typical situation, model in inference typically can be considered as post-dynamic, since they have been through the process of active construction and optimization, and thus at such point remains static for accurate prediction. Of the argument of which it can `somehow store memory', such can be said of the regime in which RNNs or LSTM, of which takes into account of the encoding, historic and compressed memory influence between passing, and further on of the same principle as TNN (Transformer). The lag and processing speed can be clearly shown with lengthened discussion, as evidence. Furthermore, the process of learning itself is external, meaning that even the basic structural sophistry of which we consider the learning agent is already insufficient with respect to current ML system and models. Instead, what we are having is in fact a toolset, of which effective in making such model to a certain degree of prediction and statistical inference, but intrinsically cannot learn, and intrinsically black-box. 

\subsubsection{Architectural Insufficiency}

Architectural insufficiency is the aspect in which we consider, the architectural design that is used in creating and constructing the construct that is evaluated for intelligence. Within its development history, AI has received numerous structural designs, some of which more successful than others. A fair share of our issues have been directed to the symbolic, logicism camp, so for now we would like to inquire on the matter of the statistical camp --- machine learning models, and deep learning framework. 

While classical frameworks have been noted of their own limitations and thereof, modern neural network structures, as deep learning architectures are much richer and much more interesting of such concern. For now, let us elaborate with the points on the architectural insufficiency observed of neural network, with respect to a provisional view toward a construct that is ideal of intelligence. 

In essence, the current neural structures are concerned of the operation of neurons together in a unit-wise fashion. This started way earlier, with precursor constructions notable as \cite{Rosenblatt1958ThePA,mcculloch_logical_1943}, and more. In modern land scape, scaling such system is often done in \textbf{layers}, of which make up of the many smaller neurons organized and sequentially activated, i.e. feedforward, where individual layers sequentially feed its outputs to the inputs of the next layer. This is usually called layer connections, in some cases, and hence if two layers are connected such that each neuron $n_{j}\in L_{2}$ is connected to every neuron $n_{i}\in L_{1}$, we say the network is \textit{fully connected}. Such operation is then repeated, and its learning behaviours controlled by the algorithm of backpropagation, of which propagate gradient errors network-wide and perform numerical correction, such that there then exists a correct solution to the given problem of the optimal configuration of the neural network. Then, at such stage, one simply fix the neural network, and let it run on the problem landscape with great accuracy. The most basic form of such architecture is the multilayer, fully-connected neural network, or multilayer perceptron (MLP), for $K$-layer, $m^{(0)}=d$ and $m^{(K)}=1$ for $m$ the width of the network, and $d$ the shape of the input vector.

\begin{definition}[Standard multilayer network, \cite{zhang2023divedeeplearning}]\label{def:SMLP}
    We define a $K$-layer fully-connected deep neural network with real-valued output. Let $m^{(0)}=d$ and $m^{(K)}=1$ for $m$ the width of the network, and $d$ the shape of the input vector. We then recursively define: 
\begin{align}
    x_{j}^{(0)} &= x_{j} \quad (j=1,\dots,m^{(0)}),\\ 
    x_{j}^{(k)} &= h\left(\sum^{m^{(k-1)}}_{j'=1} \theta_{j,j'}^{(k)}x_{j'}^{(k-1)}+ b_{j}^{(k)}\right)\quad (j=1,\dots,m^{(k)}), \quad k = 1,2,\dots,K-1\\
    f(x) & = x_{1}^{(K)} = \sum^{m^{(K-1)}}_{j=1} u_{j}x_{j}^{(K-1)}
\end{align}
where the model parameters can be represented by $w=\{[u_{j}, \theta_{j,j'}^{(k)}, b_{j}^{(k)}]: j,j',k\}$ with $m^{(k)}$ being the number of hidden units at layer $k$; $\theta\in \mathbb{R}^{m}$ the weight of the neuron.
\end{definition}

Such structural definition, of which use the unit abstraction of neuronal units, and the dynamic consideration of connections between parallel layer computational, allows for a variety of different architectural designs and application of the framework itself. Some of such includes the aforementioned neural networks in charge of revolutionary works in NLP, Graph Neural Network (\cite{Oono2020Graph,Scar04,GRP_Hamilton}) for graph-like structural data, physics ODE-based neural network like \cite{chen2019neuralordinarydifferentialequations} and so on, with numerous applications. 

The first problem that come with this type of standard architecture is the expressive problem. Standardly speaking, both classical modelling, symbolic computation, and continuous/discrete Markov chain process\footnote{Indeed, in the case of Markov process, a (discrete-time) Markov chain defines an operator \begin{equation*}
    \mathcal{T}: f(x) \to \mathbb{E}[f(X_{t+1})\mid X_{t}=x]
\end{equation*} on some space of function $f:\mathcal{X}\to\mathbb{R}$. That is, every Markov chain can be seen as defining a linear operator acting on a function space. Such operator is typically known as the Markov transitioning operator, or Koopman operator in dynamical systems.}, relies on the perspective of \textbf{function construction/approximation}, that is, we assume that, there always exists, of the best-case scenario and the internal machine assumption, the concept $c\in\mathcal{C}$ can be expressed of the function space, that is, a collection of function that can be approximated, reconstructed, and so on. This ranges from symbolic process function, transition functions, time-evolution function given a system state $S$ of variables and parameters, or simply variables-related functional relationship. As such, structures and solutions on such space relies on reconstruction and approximation on various metric $d(f,c)$ that gauges the ability to approximate the learning objective in such numerical approximation landscape. Expressibility is then required, as the question of "what kind of function class can encompass the entire concept space, and can there be a universal approximator?" for such various structures. For neural networks defined by definition~\ref{def:SMLP}, such is usually guaranteed, over certain class of functions, by the Universal Approximation Theorem, first proved by \cite{cybenko1989approximation} for sigmoidal network, \cite{hornik1989multilayer} for general activation function class, non-Weierstrass (polynomial) activation in \cite{leshno1993multilayer}, and partially, of probabilistic version on $L^{p}$ space in \cite{hornik1991approximation}. There are many statements and frameworks of such UAT is considered, though, UAT on neural network is considered in two different ways, as for two different architectural build-up of the network - arbitrary depth (layer), or arbitrary width (layer's neuron). However, in general, it states:

\begin{theorem}[Universal approximation theorem --- UAT]
    Single hidden layer $\Sigma\Pi$ feedforward networks can approximate any measurable function arbitrarily well regardless of the activation function $\Psi$, the dimension of the input space $r$, and the input space environment $\mu$. That is, for every squashing function $\Psi:R\to[0,1]$ of which is non-decreasing and $\lim_{\lambda\to-\infty}\Psi(\lambda)=0$, every $r$ and every probability measure $\mu$ on $(R^{r},B^{r})$, both $\Sigma\Pi^{r}(\Psi)$ and $\Sigma^{r}(\Psi)$ are uniformly dense on compacta in $C^{r}$ and $\rho_{\mu}$-dense in $M^{r}$. 
\end{theorem}

a perhaps simpler statement can be as followed. 

\begin{theorem}[Universal approximation theorem, simplified]
    For a class of functions $\mathcal{F}$ and a compact set $S\subset \mathbb{R}^{d}$, if for every continuous function $g$ on $S$ and for any $\epsilon>0$, there exists $f\in \mathcal{F}$ such that \begin{equation}
        ||f-g||_{\infty} = \max_{\mathbf{x}\in S} |f(\mathbf{x}-g(\mathbf{x})| \leq \epsilon
    \end{equation}
    Then, the class of functions $\mathcal{F}$ is a universal approximator of all continuous functions on $S$. We then indict that $N(w,d)$  of the neural network structure, derived from definition~\ref{def:SMLP} is such universal approximator on the set of all continuous function on $[a,b]$ of arbitrary measure $\mu$. 
\end{theorem}

While not determining correct optimal procedure that can lead to such result, or arbitrary properties of fitness, stability and so on that is useful for operational process, UAT still determines a weak \textit{existence theorem} --- such that in said domain, there exists the fundamental optimal model that can reconstruct partially the structure of the concept objective of the approximation task. Here, we still notice that we assume the observables of facts come in the form of function, and only function. Nevertheless, in particular, UAT allows for the guarantee of its expressive power over such class of function as the target. As such, many developments, of which derives from the class of optimization technique on function encoded space --- numerical encoding on $\mathbb{R}$, landscape optimization such as gradient descent, coordinate descent (\cite{luo1992convergence,tseng2001convergence,friedman2007pathwise,wright2015coordinate,tseng2009coordinate}), extension of such into the process of backpropagation (\cite{rumelhart1986learning} and older historical development). Because our argument is about architectural insufficiency, one then can ask what is the drawback in such framework? 

First and foremost, UAT and the functional structure deliberately choose and restrict itself around smooth, continuous, well-behaved function. Not taking the impossible domain lies beyond the compact subset, many functions or encoded function cannot be determined, or can be approximated partially to a given arbitrary $\epsilon>0$, yet not satisfactory. Variations of UAT usually can be considered weak, with some stronger theorems deliberately require the use of stronger assumptions and more specialized constraints.  Any given different setting, structures, data type and the encoding space of such concept, for example, as graph theoretical data, requires to be encoded or embedded into the continuous differentiable pipeline of a neural network to be utilized, and in the process making the structure suboptimal in certain standards. This result in the structure of Graph Neural Network, or GNN, of which use an encoding scheme, and a typical block aggregator over local graph information, for example, as 
\begin{equation}
\begin{aligned}
m_v^{(k)} &= \psi^{(k)}\!\left(\{\, M^{(k)}(h_v^{(k)},\, h_u^{(k)},\, e_{uv}) \mid u \in \mathcal{N}(v)\,\}\right), \\[4pt]
h_v^{(k+1)} &= U^{(k)}\!\left(h_v^{(k)},\, m_v^{(k)}\right),
\end{aligned}
\end{equation}
Where $e_{uv}$ represents the edge features, if any, $h^{(k)}_{v}$ is the embedding of node $v$ at layer $k$, $\mathcal{N}(v)$ denotes the neighbours of $v$, $M^{(k)}$ denotes the message function producing messages from neighbours, $U^{(k)}$ denotes the update function, and $\psi^{(k)}(\cdot)$ denotes the message aggregation up to layer $k$ or such node in consideration. Even though GNN of such `rigorous' in a sense framework can be used in a variety of sessions, being restricted to such neural network framework brings extraordinary drawbacks. For example, the expressive power on the native graph data is at best, equivalent to \textbf{1-dimensional Weisfeiler-Leman (1-WL) test}. Such message passing scheme often fails because of the bottleneck problem intrinsic within the framework requirement and complex backpropagation adaptation, as seen in \cite{alon2021bottleneckgraphneuralnetworks}. Further problems like structural information losses, scalability problems, trainability issues, over-smoothing, ontological structural mismatch, and so on, as seen in \cite{NEURIPS2019_bb04af0f}. Internally of the neural network framework, such problem is not restricted to such adaptation, which requires both complex restructuring and architectural mapping, but also problems like the parity or memory-sample problem in statistical query model (\cite{blum2003noise,garg2021memory}), planted clique problem (\cite{barak2016nearly}), worst-case analysis indication of neural network being NP-complete or $\exists\mathbb{R}$-complete (\cite{blum1992training,abrahamsen2021existential}) for low-level, 3-node, 2-layer configuration, computational problem and its expressibility (\cite{eldan2016power,telgarsky2016benefits,siegelmann1991turing,weiss2018practical}). Perhaps more interestingly, is the actuated and controversial No Free Lunch theorem from \cite{wolpert1997nofree}, of which basically states that a "structural-less", or naively defined structural copulative information, construction on the encoding scheme, is equally bad as with any given model of arbitrary sense, and also the learning procedure associated with such. Setting this up with respect to a particular set of preset inductive bias, or structural assumptions thereof, scaling exorbitantly both of components, parameters, and repeated structures like typical deep learning structure, is not feasible and reduce its success to uncertainty, as the stronger, relative-NFL theorem takes effect --- given arbitrary structural information as the origin, there then exists, particularly, certain point of which without further information introduced to the setting, the supposed `information' can be reiterated as misleading, thus average out the model's evaluation and performance. 

Aside from some archetype of learning, for example, online learning of which keep dynamically update the model in a sense, generally, neural network and such type of parameterized models are \textit{static}. They act, just as CRA indicted, to be an algorithmic machine only, of which fits to a description, to a task, to a given degree of accuracy within certain manually designed encoded space, manually designed objective, limited or no interpretation of the scheme itself (classification machines think in numerics and matching errors, thus would not know what a cat is, but only know that an object named \texttt{0x0AB4} must be evaluated in some way that can be optimized and reduced). The operating stream itself of the neural network is limited, as we can say of the feed-forward scheme. The extension that neural network is configured to is limited, and lest to say that the neural network scheme itself is underdeveloped, ultimately relies on the simplest possible modelling structure to then give rise to a multitude of different architectures of date. Computational optimization and layer-optimization, practically constrained structures further hampers down such understanding of the neural network itself, as nobody knows what happen in the neural network, less to try adapting it to different situations. 

Go for technicality, neural network architecture of itself faces several increasingly difficult dilemma. Let us disregard the problem of Neural Scaling Law, \cite{kaplan2020scalinglawsneurallanguage}, as it is simply an empirical optimization observation hyped of its wording\footnote{Of the Neural Scaling Law, conceptually, the result is purely empirical on deviation from regular norm, which is expressed by $$L(N,D,C)\propto (N^{\alpha}D^{\beta}C^{\gamma})^{-\eta}$$ Where $L$ is the loss, $N$ being model parameters, $D$ is the dataset size, $C$ being the computation used, and $\eta$ is an empirically fitted exponent. Plotting such relation into logarithmic scale gives roughly the linear relation that is typically seen in literatures and news context. The apparent `law' only emerges because it is plotted on log-log axes. Almost any monotonically diminishing function can be made to look linear on a log plot over a limited range. The law itself also only empirically states such relation without the underlying mechanics or assumptions that would make it a technical `law', and is purely empirical observation. In practice, achieving near-zero error is fairly normal, and within such structure, there exists no framework to indict that the neural scaling law holds any value aside from interesting engineering heuristics. Indeed, perhaps, after it is published, many papers were born dedicated to question or beating such argument, such as \cite{sorscher2023neuralscalinglawsbeating,ivgi-etal-2022-scaling,su2024unravelingmysteryscalinglaws,lee2025neuralscalinglawsleading}, and \cite{doi:10.1073/pnas.2311878121}. Therefore, it is not worth it of such analysis.}, the theoretical ground on which neural network is born from, and the empirical method it employs to advance ever since, face problems regarding interpretability, structural effectiveness or definition, phenomena explanations, uncontrollable behaviours, black-box restrictions, rigidness of architecture, uncertainty and vanishing problems, overloading problems, and so on. Interpretation is the largest problem with such theoretical and both heuristic treatment, as no one understand what lies underneath such system and the operation that births such result observed or tasked itself. Explainable AI, like presented in \cite{hsieh2024comprehensiveguideexplainableai}, was created to counter such issue, still cannot work or cannot make substantial advancements aside from careful manual designs and limited domain analysis, which in an unfair bit of comparison, go back to the time of symbolic AI approach. This is even worse for neural network, as the regular wisdom is that no one understands the operation of hidden layers and all, in any given setting, aside from which you reduce it to very small size like a perceptron or so. Theoretical development toward such is also slow, and often simplify it to mathematical object to be analysed too rigorously, losing the essence of the architecture in the mathematics rigours. This is particular event or rather pattern that we unfortunately stretched to the previous section on learning theory. Often see in learning theory development, or machine learning theory in general - the `\textbf{mathification}' of a theory is a problem that somewhat plagues papers and researches on machine learning topics. Even though we need mathematics on either end, the approach is impractical or simply wrong, believing mathematics to be the singular thing that defines learning theory. This has many fallacies that can then be attributed to a lot of factors and whatnot, of several factors that plague this analysis even further than just the problem of double descent. For now, 
\begin{enumerate}[itemsep=2pt,topsep=1pt,leftmargin=1.2cm]
    \item Epistemological limits and interpretations (\cite{electronics13020416}) - We ultimately lack understanding of a lot of things. While those `theorems' are very nice in learning theory, the real picture is that it is not real learning, for the word learning are not even defined, as such is to compare them to \textit{human learning} on itself. Even by then, theorems are severely limited. Certain voices also concern of similar problem, including \cite{lipton2016mythos,doshi2017towards,molnar2020general} (\cite{molnar2020general} shortly consider the misleading interpretation question instead), and philosophically, with several pushbacks on structural anecdotes, \cite{dreyfus1965alchemy,dreyfus1972what,dreyfus1986mind,suchman1987plans,brooks1991intelligence,searle1980minds,mccarthy1969philosophical,harnad1990symbol}. On the more modern side, of contemporary critiques, \cite{pearl2009causality,marcus2018deep,sutton2019bitter}. Of the Chinese Room Argument, perhaps we can look into already copulated passages. 
    \item Applying wrongly, and is used to impress and not to explain anything (\cite{lipton2018troublingtrendsmachinelearning}) - mathematics is used to \textit{impress} certain demographic of reviewers and readers, to provide a sense of rigours, to further enhancing the image of formal theory to the point that such theory, even if wrong, can be considered fairly correct by the sheer volume of practitioners believing in such. Such is to say the \textit{mathiness} is turning things into ideology more than rigours itself, of which we can take a tangent to see in economic theory, one of the place to adopt a large portion of machine learning statistics, the pushback against such cursed devolution (\cite{romer2015mathiness,syll2024postreal}). Such can also be seen in generally most double descent analysis, in which there exists many formalisms yet not definite result on double descent. 
    \item Reproducibility and acute false claims (\cite{kapoor2022leakage}) - In general, what we have done cannot be recreated, in certain way, and of certain too optimistic setting that is seemingly unrealistic - particularly in adoption toward practical means of actions. Certain theoretical assumptions and formulations are too strict, of which means the increment of hypothesis fluctuation with removal of such constraints. 
\end{enumerate}

Continuing, practical problems range from grokking (\cite{power2022grokking,davies_unifying_2023}), double descent (\cite{belkin_reconciling_2019,nakkiran_deep_2019}), triple descent (\cite{d_ascoli_triple_2020}), counter-intuitive results (\cite{szegedy2013intriguing}), adversarial examples or statistical instability (\cite{goodfellow2014explaining,carlini2017evaluatingrobustnessneuralnetworks}), `memorization issue' (\cite{arpit2017closer}), generalization problem and its purposes (\cite{zhang2017understandingdeeplearningrequires}), catastrophic forgetting (\cite{kirkpatrick2017overcoming}), and so on. Terms like inductive bias are thrown in of many contexts without a single grounded definition or notion to support such, many terms where born without actual consideration, concepts are thrown in and out without basis. While it is easy to simply ignore such problems, such as to be seen in usual deep learning practices of AI, the crack inevitably shows, and a very wide range of developmental gap ensues. 

\subsection{The end goal of AGI}

With that said, what is the end goal of AGI? In all of it, the central goal is supposed to be the creation, out from the doctrine of fragmented intelligence, a unification in which we can then centralize everything in a singular artificial intelligence system, hence AGI. However, as we stated, the term AGI can be further dissected to subsets of such AGI construct, and even then, different system has different AGI threshold. Thereby, the end goal is not clarified yet. What we are seeing instead is a wave of hysteria, from people afraid of the incremental development, and the current market bubble generated from the present AI boom, all of which in turn, pushes the notion of `building AGI' further and further to the truth. Such is also amplified of the outlook to the concept of ASI, of which is sold in the general media as the next stage on which post-scarcity can be achieved, where Nobel-level researches are outputted every few days, and automation leaves human of no burdensome tasks. However, with what is being done right now, it seems like such outlook cannot be realized, at least in due time. 

The conflict in which the philosophy of which promised AGI to the public, as well as the optimistic outlook of people and thinkers, perhaps can be reflected in the sense of Heideggerian philosophy, or famously so, the philosophy between \textbf{ready-to-hand} (Zuhandenheit) and \textbf{present-at-hand} (Vorhandenheit) (\cite{heidegger1962being,dreyfus1991being}). Heidegger simply explains, of the notion in which the object appears in front of us. In a sense, the subject of question, and the person of inspection, of all but anthropocentric. For Heidegger, most of the time we are involved in the world in an ordinary way or "ready-to-hand." We are usually doing things with a view to achieving something, and hence, with a purpose already configured, and project such voice outward. The being of the ready-to-hand announces itself as a field of equipment to be put to use, and hence ultimately defines itself around such action and potential. A famous example of such notion is the hammer analogy. In question toward what the hammer can do, whether because from intuition or else, we extrapolate such object of its functionality, instead of inclining on the more sophistry of decompositions and analysing components. Such knowledge can then be made trivial, or thus transferred to a lower priority of thought, of which to the point where one can talk comfortably of other subjects or adjacent topics, but do not require careful contemplation to the original consideration (the hammer). The reversal of such action happens in isolation, in a typical understanding of which decomposition is required, structural knowledge needs to be formed, and so forth, such is to say we purposefully study such object similar to a scientist in idealization studies carefully of certain object, and hence, defers its meaning and existence to the notion of it "exist" to be there. That is the idea of present-at-hand.

It is then fairly connective to approach from certain angle, that concerning Heidegger's notion, our AGI system, plus the non-elucidating way of studying the architecture of choice for such endeavour of AGI, we deliberately refuse the notion of present-at-hand for a more ready-to-hand, in some cases to the extreme of \textit{empiricism}, of which then also founded the principle of agentic AI system. This can be applied to even further, as to the normal principle and conceptual architecture of artificial intelligence system in an agent-environment scheme as being inadequate, correct, but not enough (\cite{stanford2018ai}). There will be, ultimately, a wall in which this approach cannot reach, as to there exists too many permutations of a path such that there exists fundamentally, of a statistical approach, to be an almost zero chance in which such can happen, within the wrong mode of understanding, both from the perspective of the designer, and the perspective of the construct itself. In the same spirit of the UAT theorem, we can say that the optimism will be, to a certain point, similar to saying "Somewhere in the space of all possible English sentences, one perfectly describes quantum gravity.". Interesting and true (probably), but useless.

\clearpage

\section{Future of Artificial Intelligence}\label{sec:3form}

The future of artificial intelligence under the current framework looks bleak. Indeed, such can be said of every `matured' enough research and resource-intensive master research plan, and thus, we would have to consider an almost likely slowdown and perhaps not so dramatically, a winter age again for adoptions and maturity in other facet of technological application. During such time, however, we might as well look at what to do to try resolving such problem. 

\subsection{Neural architecture formalism}

We would argue that the neural network idea is indeed, formal and foundational, more than it is usually attributed of. Formally, a neural architecture follows inspiration from the biological neuron in the human brain, and thus employ the philosophy of unit-based processing (\cite{Rosenblatt1958ThePA,10.5555/50066,mcculloch_logical_1943}). This implies various properties. First, they are categorized conceptually into units of processing, of which all neuron $n$ admits the structure $(I,M,O)$, of the input receiver $I$, the internal processor $M$, and the output transceiver $O$. Any neuron admitting this structure then can be constructed, combined, and fully realized at will. Conceptually, this created the \textit{typed} of the neuron architecture, in which all structures would have the same type for operations, such as composition of two neurons, $n_{1}\circ n_{2}=n_{2}(n_{1})$, in this case connected to each other by sequential move, for all neuron to have the sequential operation $n_{i}: I_{i}\times M_{i}\to O_{i}$. In practice, handling this might require more careful planning, but nevertheless the structure is particularly streamlined. Another property that is implied is the recursive structure that can be employed. Intrinsically, the admission of $(I,M,O)$ structure implies that any given structure can mutate $M$ for different purposes, for different processing and thereof, as long as $I$ and $O$ stays as immovable component of the neural structure. Then, Any nested sequence of neuron can be compressed to be a singular neuron, since the footprint of the operation $n_{1}\circ n_{2}$ is simply 
\begin{equation}
    n_{1}\circ n_{2}\equiv n_{3}: (I_{1},M_{1})\to O_{1} \to (I_{2},M_{2})\to O_{2} \equiv n_{3}: (I_{1},M_{3}(M_{1},M_{2}))\to O_{2} 
\end{equation}
in which we clarify $M_{3}$ as the processing equivalent of both the first and the second neuron. Thus, we can nest many structures together, changing dynamics altogether, and create different type of specialized neurons, yet with only careful planning of the pair $(I,O)$ we can operate on them together at will. In fact, one can simply also connect as many input from $n_{1}$ to $n_{2}$, neglecting the rest, and the neural structure will still work. We say that $(I,O)$ represents the \textbf{signature} of the neural structure. A baseline, minimal neuron structure can then be defined, which fits the basic definition of a singular perceptron in theory. Let us define $\mathcal{N}_{i}$ as the $i$th arbitrary classification of neuron class. We define the criteria of minimization as

\begin{definition}[Minimization set]
    Let $x$ be a neuron of arbitrary neuronal classification $\mathcal{N}$. Then, the requirement of all neuron class is to be able to distinguish its component to three parts, that is, $\min_{\mathcal{N}_{i}\in \mathcal{N}}{\mathcal{N}_{i}}\equiv \mathcal{I}, \mathcal{M},\mathcal{O}$ where $\mathcal{I}$ is the input channel, $\mathcal{M}$ the internal mechanics, and the output $\mathcal{O}$. Let $i,j,k$ represents the cardinality of each part respectively, then if
    \begin{equation}
        i=j=k=1, \quad \min_{\mathcal{N}_{q}\in \mathcal{N}}{\mathcal{N}_{q}} = \mathcal{N}_{q}, \forall q \geq 0
    \end{equation}
    Then we call this class of neuron the \textbf{minimal neuron class}, and any $x\in \mathcal{N}_{i}$ of such is called the \textbf{minimal neuron} or \textbf{standard neuron}, denoted by $x_{S}$. By default, this is satisfied if $q=0$ in our construction. \footnote{The constant $i$ here refers to the organization numbering of nested classes built upon by another components. In such, we observe that this construction implicitly defines itself to be the simple zeroth class.}
\end{definition}

Then, we defined the class of all minimal perceptron $\mathcal{N}_{0}$, or neuron unit, as followed.

\begin{definition}[Class $\mathcal{N}_{0}$ on $\mathbb{R}$]
    A neuron unit $x\in \mathcal{N}$ belongs to class $\mathcal{N}_{0}(\mathbb{R})$ and is called a \textbf{standard neuron on $\mathbb{R}$} if it satisfies the minimization set criteria, and can be written of the form: 
    \begin{equation}
        x = q = \sigma_{\mathcal{M}}(w\cdot p + b),\quad  p\in \mathcal{I}\subset \mathbb{R}, w,b \subset \mathbb{R}\subset \mathcal{M}, \sigma: \mathbb{R}\to \mathbb{R} \in \mathcal{M}, q \in \mathcal{O}
    \end{equation}
    If $\sigma$ is linear unit, that is, $\sigma(wp+b)=wp+b$, then we say $x$ is a \textbf{linear standard unit}. \footnote{One might ask why we use the product and addition in the formula of $wp$ and then $b$. In fact, this is perhaps more trivial - as to facilitate the concept of \textit{linearity} - the formulation looks exactly like the linear line in a plane. Furthermore, as we will soon see, it is also of interest such that units of neurons can be linearly combined in a way, at least of computational aspect in running it on computers.} 
\end{definition}

Naturally, a singular neuron is not enough, and as illustrated in \cite{10.5555/50066}, they alone cannot do everything, for example, the XOR problem illustrated that particularly, there are unsolvable problems one can get with a simple perceptron. The resolution to this problem come in form of, as implied, of the streamlined nature of neural structures --- what if we operate them in parallel, in larger structures called \textbf{layers}, and so on? This is first illustrated by Rosenblatt, and its elementary form varies a lot in the history of classical connectionism. By the form of neuron class, we classify it $\mathcal{N}_{2}$. We reserve class $\mathcal{N}_{1}$ for the class of all \textit{multiple-input neuron}, of which the cardinality is $(i,1,1)$ for $i=1,\dots,n$. The motivation for $\mathcal{N}_{1}$ is that to resolve the problem of the class $\mathcal{N}_{0}$, one potential fix would be to 'fix bayonet' and free up $i$, thus giving the construction of $(i,j,1)$. We call this \textbf{multivariate neuron}. If it is $(i,1,1)$, then we call it the \textbf{multivariate standard neuron}. All of such neurons then belong to the \textbf{class $\mathcal{N}_{1}$ neuron} simplex. Then, the class $\mathcal{N}_{2}$ of layer neural networks, is defined as followed. 

\begin{definition}[Class $\mathcal{N}_{2}$ structure]
    We fix the signature of any given structure $\mathbf{N}\in\mathcal{N}_{2}$. Let us define, for $L_{i}\in M$ the structures of layers, of which $L_{i}$ contains $n_{i,j}\in \{\mathcal{N}_{0},\mathcal{N}_{1}\}$ of subsequent lower class, and fix their cardinality of the form $(i,1,k)$. Then, a \textbf{neural network} $\mathbf{N}$ of the class $\mathcal{N}_{2}$ is equivalent to the following structure: 
    \begin{equation}
        \mathbf{N}\in \mathcal{N}_{2}\equiv I_{\mathbf{N}} \times M_{\mathbf{N}}(L_{1}\times L_{2}\times L_{3}\times \dots \times L_{j}) \to O_{\mathbf{N}}
    \end{equation}
    The cardinality of $\mathbf{N}$ is then $(i,j,k)$, for all $i,j,k\in\{1,\dots,n\}$. For $\mathbf{N}(i,1,k)$, we call it as the shallow neural network. 
\end{definition}

The structure of our theory on the neural formalism is influenced by the object-abstracted treatment of mathematically embedded structures, and the unit-wise principle of particular neuron. Before we meet ourselves into the notion of epistemic circularity\footnote{Also called bootstrapping, where to understand or define certain notion, requires the knowledge of the object wishes to be defined itself - for example, defining the size of a finite natural number set, using the elements of the number set itself.} problem, we might as well clarify a few prerequisites for such structure to exhibit. 

First, we indict on the fundamental encoding environment that any object can take. The main point of any structure here is that there exists fundamentally the encoding space of two types. First is the object's cardinality space, denoted $\Gamma=(\mathbb{N},F)$ for any given categorization $F$. Second is the encoding primitive of the field $\mathbb{R}$ for generality - in general any field is alright, and they define the analytic structure of the system itself. Any extension, for example, the $\mathbb{R}$-algebra of complex number $\mathbb{C}$ is then the primitive field's extension. \footnote{This corresponds to a variety of ideas. For example, see \cite{Cartuyvels_2021} for the idea of discrete and continuous representations and processing, and \cite{M_ller_2022} for the same idea, but used on computer graphic process where it is constructed as discrete indexing structure (hash tables) with continuous neural fields. Cardinality space takes inspiration from combinatorial species, sheaf theory of global-local set, and Grothendieck's approach to algebraic topology. Extension is the idea from field extension itself, for structures that can be extended. The general idea of this section relies on interpretation of category theory.}. Any type of data or system can then be decomposed to such, with additional structure on top of such primitive. Such is then called the \textit{primitive framework}. 

\begin{definition}[Primitive framework]
    Let us define the primitive framework $\mathcal{P}_{0}$ of the dual $(\Gamma, \mathbb{R})$ where $\Gamma=(\mathbb{N},F)$ is the cardinality encoding space, and $\mathbb{R}$ is the base field primitive of the analytical encoding. An object $X\in \mathcal{P}_{0}$ admits a dual representation, 
    \begin{equation}
        X \mapsto (\gamma(X),\rho(X))
    \end{equation}
    for $\gamma: \mathrm{Obj}\to \Gamma$ of cardinality encoding, and $\rho:\mathrm{Obj}\to\mathcal{E}(\mathbb{R})$ for the field extension of $\mathbb{R}$ category, or the category of all $\mathbb{R}$-algebras. For $\mathcal{E}(\mathbb{R})$ without extension, then $\mathcal{E}(\mathbb{R})\cong \{\mathbb{R}\}$ of all $\mathbb{R}$-algebra. 
\end{definition}

For a structure that is supposed to be unit-wise constructed like neural network and the like, one of the main principle is the principle of abstraction. With this, come the idea of layer. Specifically, a \textit{layer} separates abstraction in terms of subspace. Let us take an example of such kind for clarification. Let us define the primitive framework $\mathcal{P}_{0}$ as now the layer $L_{0}$ of this layering scheme. Then, we define the layer $\mathcal{L}_{1}$ of all unit-wised neuron-like units taking over the representation scheme on $\mathcal{L}_{0}$. We then define the structure of the neuron class $\mathcal{N}_{1}$ upon such as followed. 

\begin{definition}[Base neuron class]
    We define the base neuron class $\mathcal{N}_{1}\equiv L_{1}$ as followed. For any $\mathcal{U}\in L_{1}$ for $\mathcal{U}$ as unit-wise construction over $L_{0}$, then every $\mathcal{U}$ satisfies the input signature $\mathcal{I}_{sig}:\mathcal{E}'(\mathbb{R})\to L_{0}$, output signature $\mathcal{O}_{sig}:L_{0}\to\mathcal{E}'(\mathbb{R})$, and $\mathcal{C}_{ext}$ as extensible construction over $L_{0}$, for $\mathcal{E}'(\mathbb{R})$ particular extension on analytical encoding of $L_{0}$. The \textbf{type} template of a neural unit $\mathcal{U}\in L_{0}\equiv \mathcal{N}_{1}$ is then defined as $\mathcal{U}=\langle \mathcal{I}_{sig},\mathcal{C}_{ext},\mathcal{O}_{sig}\rangle$ where $\mathcal{I}_{sig}$ and $\mathcal{O}_{sig}$ are invariant as type. 
\end{definition}

Those definitions directly link it to type theory, while such development is perhaps more complex. In general, this specifies the \textit{type}, of a particular unit using the invariant analytical typing in the general environment linking to it. This is the \textit{outer typing} of a given model construct, of which defers the type of which interaction between the environment, or the \textit{global space state} happens and what is received of such. While the ambiance space can be forgiving, such cannot be said for the typing of given structure, since it must have the correct typing for identification. The extension is then is fairly simple, but the more important notion is the equivalence of the variant $C_{ext}$ extensible unit of the inner structure itself. 

Another type of extension can also happen in the sense of $\gamma(X)$. In certain sense, $\gamma(\cdot)$ can always be richer than the cardinality encoding, but with the use of extensible precursor-usage representation. Suppose a grid-like ordering can be made of the spatial ordering type using information of the precursor $\gamma(X)$. Then, such object then is extended to be a \textit{dual-extended representation}: 
\begin{equation}
    X\mapsto (\gamma(X),\rho(X))\times F_{\gamma}(X)
    \label{eq:extend_gamma}
\end{equation}
where $F_{\gamma}(X)$ organizes the different constituent part in cardinality argument, of positional placement, or else. Such freedom can be made and thus, categorize different extensions into various packages. This can then be extended further and further of such, thus making it more sophisticated and will reach, fundamentally, to the current neural network, its specialization and variations, and so on. Indeed, we have implicitly obtained the hierarchical construction
\begin{equation}
    \mathcal{P}_{0} \subset \mathcal{P}_{1} \subset \dots \mathcal{P}_{n}
\end{equation}
with increased specific direction of increment. This framework a notion of which in general, can be called the \textit{specific constraint chain}, in which a specific toolchain of increment recursive construction is created, from a transformation $\Phi_{i}:\mathcal{P}_{i}\to\mathcal{P}_{i+1}$, thus create a tower of hierarchy of framework. With any self-component extension, for example, at the layer $\mathcal{P}_{L}$ (of which in certain construction as $\mathcal{P}_{2}$), we can define the add-on $\Lambda_{i,X}:\mathcal{P}_{i}\to\mathcal{P}_{i}'$, that modifies the internal structure with added mass or added cardinality, but retain the core principle construct. 

While this is a prototype, it certainly focuses on the main axis of development that would be detrimental to the topic of \textit{interpretability}, capacity, analysis and so on. We can indeed, extend this particular base template of a neuron class to much greater strength, of which address partially concerns in both implementation, and analysis. Type and neuron specification classifies neuron into different \textit{class} of components instead of the mundane functional structure that we currently have, but also individualized and structuralized the \textit{connection, information} and signature aspect of a processing unit. This allows for more grounded expression and composition rule, for example, the complexity metric of individual components of a larger construction of neural network, of which then can be expressed by individual neurons, and such neuron's individual's components. Dynamics works the same, however, it will shift - instead of learning as a back-tracking process, we can express backpropagation as a kind of \textit{internal mechanism} on addition to the working framework, and the components such measures. In all, the framework is not category per se, but a layered analytic machinery of which naturally (or at least intended of) to construct and describe how architecture can emerge from iterated constraints extension of primitive computational units. In essence, we can also treat the framework as to say each $\mathcal{P}_{i}$ to correspond to a functional space of \textit{realizable computation} (as seen in the abstraction to $\mathbb{R}$-algebra at least), and different structural addition. Furthermore, one can also use such framework to either construct from scratch, and also enables of the process for structural optimization, basically, turning the system considered in such framework, into a basis, resource-optimization or construction game, where analytics and structural design are grouped and unified, and also taken into consideration more sophisticatedly as a resource. Such submodules' implication can then also contain the \textbf{learning mechanism module}, of which now is intrinsic of the system, and not simply external of a process itself. 

Furthermore, the primitive framework itself serves of a detrimental role in expressing system-wide, modelling-style and global specification of the working space. While we speak of hypothesis class $\mathcal{H}$ or $\mathcal{C}$ for example, such class can only infer, usually of specific function class, definitions, requirements, and so on. With this, we can express them both in a set-theoretical way of cardinality and field-extension similar to embedded vector space, and also encoding such as the numerical encoding typically seen in computing system, and structural addition and expression explicitly considered so. The goal is then also to provide a concrete groundwork on conceptually model the system dynamics and its constituent parts, in the face of increasing developmental pace and new architectures floating around from heuristic choices. In similar sense, using only category theory is an ill-advised method to formalize such, and hence we would like to not agree on such stance. Indeed, there are problems and pitfalls, as well as the immaturity of the functional structure itself, but per a prototype, the general conceptual idea of this particular framework is fairly sound, and would be able to provide deeper insight and more sophisticated interpretation, rather than just simple neural network functions before. 

Again, such theory and treatment does not claim to solve the problem directly, as even now it is not elucidated yet if such model can be even feasible of holding certain conceptual works. Instead, what it might do is to extend and allow for the backward construction - by formalize in the sense of restructure the current messy architecture, and enable the incremental construction in both depth and breath of the neural unit concept. If such can be enabled, emergence, percolation, and so on, could be reached in due time. It is then the task to develop such theory toward maturity, while still retain backward compatibility toward the simpler and present system. 
\subsection{The learning theoretic}

The most important theory or rather, functionality of mutation that enables certain model to facilitate dynamic changes, are indeed the theoretical notion of the learning procedure. Toward such end, generalization of the theory itself, and what is meant of the notion that was typically seen, for example, \textit{generalization}, and so on, is in the question for clarification, and both development of such. This section will only outline the basic philosophical thought of the implication of a learning action in the machine-theoretic lens, for further expansion of the topic to be reserved in future works and separated developments. The main goal, however, would be to realize the entire picture of how the learning process is expressed, usually, and what can be gained from such framing. 

What we attempted as learning would be inherently built upon the perspective of a system dynamic. Specifically, we see that we can utilize existing idea, for example, the simple \textit{agent} system in \cite{stanford2018ai} as the critique system. The implication is then simple - we posit that within this observation or insight, we can reveal more about the dynamics of which our models are taking, and what to construct next of requirements. For now, let us denote the agent of interest as $A$. This agent has the internal mechanism $M$, the sensory $S$ and the action $F$, of which it uses of the operating environment $E$ surrounding of which $S$ can sense, marginally of its intended features. We posit the following definition, of which applies for all constructs, lest of only the singular artificial intelligence subject. 

\begin{definition}[Construct]
    A \textbf{construct} is a conceptual encapsulation of two components: The \textbf{machine} in broad term which houses the \textit{operational facility}, and optionally, the \textit{existential facility} of the construct, and the \textbf{process}, or rather, its \textbf{state(s)} of being. These two makes up a functional construct of interest.
\end{definition}

The existence of such model can be realized in certain `situation' by the resource it occupies of the existence of such. Under the computationalism approach, such includes \textit{time complexity}, memory allocation, and \textit{structural interpretation} via the native \textit{encoding language} (such is numerical). The observation system of interest includes the resource as the \textit{observations} itself - usually expressed also via the native encoding language itself, attaining the structure choice of the environment - for example, continuous observations or discrete observations that can be isolated of a concrete state. Such observation itself, is governed by the selective rule of the system's environment, of which draw out such result we see of the environment. Those \textit{natural information} are inherently hidden, from the perspective of the agent sensory system itself, or rather of the inherent way of receiving information from the sensory unit configured of choice. Moving on from such, we emphasize the need to clearing out the problem of \textit{existential facility}, and \textit{operational facility} of any given agent system of interest, and thus, also the environment at will. While it can be said that the environment would be less susceptible to such framing, from the perspective of an objective constructivism approach, the environment itself being used on said agent, is within the limit in which the designer (human, etc, intelligent agency), can argue and consider of such. Thereby, there exists existential facilities - those that support such construction and regard of the environment (for example, a physical environment consideration requires the basis of laws of physics for potential framing of such laws) that is procured to the model itself, and thus form the basis of the existential facility thereof. Operational facility comes in hand as the actual functional, observables that exist as object of interest within said environment $E$. The model itself attains such level of existential-operational facilities on itself, however, the existential facilities here form the basis of its construction, while the operational facilities form the basis of its actual process and operations thereof itself. Simply speak, such notion of more naturally applied for an \textit{actionable}, \textit{constructible} system of interest. For the agent itself, computationalism induces on the property of being constructed on the basis of the computable system, i.e. computers. Thus, it bears constraints, encoding specification, structural mechanism of the computable system that it is based on upon. For example, this can be as we have already mentioned, the need for \textbf{numerical encoding} in modern computing system, the memory and time-constraint of which makes certain operation or specific operation of interest infeasible of the resource sensitive to such computational object ($NP$-hard problems, etc), or the complication in constructing certain architecture or expression of the model agent of interest (for example, a system of language, abstracted to either procedural or OOP, presents architectural abstraction in which can make constructions difficult even within a singular basis of computation). Such makes up how and what can the paradigm, for even the biological neuron abstraction can take within the respect of such computational system, reduced to an input-and-activation unit, or later on where the more computably-optimized schema is the layering scheme. Such is then can be linked to the operational facilities, of which contains two separate entities of consideration - the mass, or rather, \textbf{facilities}, and the \textbf{process}, of which considers the operational itself of the agent. 

Inherently, we see there exists two internal mechanisms of interest. One, belongs to the model itself, now we denote $M_{A}$, and one is the mechanism, or the law of which observations in $E$ is observed, denoted $M_{E}$. Such internal mechanisms are often more considerately important than the sensory actions itself, and of which generally comes in two main categorization. The internal mechanism classified via exposure, contains the \textbf{internal interface} of which is more closely linked to the specificity of the sensory system and external action potential (i.e. the action unit), and the follow-up fully internal system of which can be theorized to control and operate in unification of such sense. In a sense, what we define is the delegation of interface, in which the existential and operating facilities are handled via separated specialized interface, and so on. While it is said so as to in one way or another, partially support the theory of fragmented intelligence, such is also needed to be reminded of that in terms of internal structure, such interface bears heavy deviation toward the sensory experience itself. Or rather, the action handling is inherent of the dynamic observed by and acted upon by the sensor, and usually do not reflect such of the internal structure, aside from the internal structure's interface strength, and the effectiveness of the interface-to-sensor structure. When speaking such system, $M_{E}$ is delegated observations in which $M_{E}\to M_{E,S}$, or $M_{E}'$ for short, taking into account the sensory perception itself. The \textit{information} received from $M_{E}$ by $A$ is inherently shallow, and can only be interpreted inside $S$'s capability itself. However, such information is extended in a different case. 

Next, we further the position that there exists the \textbf{process} of the operational facilities, of which handles the operation itself. This concept requires us to handle the concept of \textbf{state}, as per natural it is of to classify and quantize the `condition' and `resource', or `structure' of a given subject, model, theory, and so on, at any given situation, without mutation\footnote{The concept is again recursive, as the `state' of a theory requires the theory of state, and so on so forth. Even for abstract mathematics, talking about the `state' of the field at any given time means to quantize the field itself, in one way or another befit of the alma matter, and thus the condition of the field as said.}. Such discrete state is handled and intrinsic of any operating structure, hence $M_{E}$ and $M_{A}$. While we do not go into many details of the philosophical inquiry about state transition, one thing can be sure is that they hold more information than the simple static observation. Thus comes the notion of a \textit{static} system observation, or \textit{dynamic} system observation. The singular $M_{E}\to M_{E,S}$ represents the static case, in which observations are frozen in time or any arbitrary notion $t$ as a `snapshot', while the copula $\{M_{E}\to M_{E,S}\}_{t}$ represents the dynamic case, in which such arbitrary `state snapshot' now contains the relative previous or future states of the system. Hence, it allows for multi-facet information, with respect to the static case of such. The system can have these properties in many ways, for example, if the system itself is a mask, then it contains the notion for the static-dynamic mask only, while the lower-level under that mask can still be dynamic - thus those concepts are system-dependent. 

What we have said are to set up the system by itself, the model in which operations are there to work, and the system observation in which $A$ itself are tasked of, given its capability thereof that can be quantified. Now, come to the question: what is the learning dynamics, and how should it be important of? 

Naively observing such, we can understand that the \textit{sensor} itself is not good of forming, acting purely as the information receiver themselves. Without the interface, it can do nothing, but we also encounter such fact that without any knowledge or interpretation, the interface only serves as an increased per length tool - just like a bridge from $Q\to P$, it does not do anything beside lengthens the signal travel time. Thereby, it requires interpretation to be built upon, in which development of the interface on sensors must be indicted of. It is also here that we see the limitation of the sensory information received from $M_{E}$, and the limitation coming from $S$ itself. $S$, even in dynamic of static case, cannot see everything there is of the measure provided on the landscape of the system. For example, if the system expresses itself majorly in $(x,y,z,t_{1},\theta)$, but the sensor can only connect to, or know of, $\theta$, this leaves an extreme gap in information. Assuming such snapshot copula for $\theta$ alone to be influenced by a network in which the other quantities interact, with its own underlying constructs and interactions, then abnormality, sudden changes, outliers, roughness of $\theta$ cannot be explained any further but to set it as noise. The snapshot itself might also leave questionable gaps between observation itself. For example, there might only exist the state in which energies are there in discrete intervals, and there are regions in which no observation are made, mainly out of \textit{structural constraints} on the system of its value on itself. The snapshot itself also does not hold any information, without interpreter, even in dynamic case --- as for in such case can only be no more than a bunch of discrete and individual observables without structural adherence, i.e. we see it as function, but the system itself experience it in a non-function sense, would not be able to triangulate such. This is fairly mitigated when the data itself is rich enough in information, and thus allows for what is usually called \textit{single-shot operation}, however, such case is typically rare, and within the current framework, the operation that led to such development is more likely to be categorized as inconsistent. Thus, we can then consider it a disorganized bunch of observation, and not inherently obtains the notion of relation, let alone function. Such pushes for the internal mechanism, or any supporting substrate, that can `interpret', using our analogy, and construct certain system inside the modelled agent itself, both in restricted case of singular subject, or in dynamic cases of data, and within poor or rich data quality and observation density. 

This, naturally expresses the notion of \textbf{learning} as an integral part of the system dynamic, however much more sophisticated in the sense of arbitrary 'adaptation' of a system instead. From such, we can clearly distinguish two types of learning - one of which is \textbf{structural learning}, where the adaptation availability is for the structure of the model itself, while the other is of \textbf{operational learning}, where the process itself is modified of said adaptation. They allow for different mode of mutation, and also different mode of modification with specified behaviours. The learning sequence then can be copulated into different parts, of which we essentially posit that structural learning is of higher priority than operational learning, though in certain sense, operational can be utilized much better. What that said, we need to take on the priori on which learning is necessary. While we have been speaking of the necessity of it as a functional construct to interpret, create interface of which we can apply on, and so on. But we are skipping the natural justification of learning mechanics, which is very hard and is not relevant in our creation (though might be so in a difference sense\footnote{Some might notice this as the point of \textit{emergence}, not in the sense of pop-science emergence, but percolation/emergence, where the property in which simple system are coupled together, working in different scale, conditions, and so on, creates complex and often reactively sophisticated system with certain amount of functionality, purposes, dialect, or simply population for discrete replica, and they themselves are parts of a larger construct interacting in such system space.}), we need to resolve, then, the \textbf{objective} of learning. This is a very hard question, and indeed, have no satisfactory answer. Indeed, the only way that we can typically see, is the scheme in which we introduce \textit{logical impulse} in which is interpreted of a sense, as favourable. That is, clarify an objective $\ell$ to reach, then of certainty, introduce the reward/punishment in which either get you closer to the objective, or further out. This hits the wall of replication and mimicking, but no actual constraints, as we have argued in previous sections. Typically, what we do in both supervised learning or semi-supervised learning scheme, or even unsupervised scheme, we often resort to either 'specialized designing' - in the case of unsupervised learning and all, or \textbf{external modification} as of supervised or semi-supervised case, in which as said, remove the learning functional, outside the working model itself. Inductive bias - or so is what it is called of, as the bias from the designer perspective, coming to the model, and so on so forth. External modification relieves the model from the burden of determining the meaning in its own space, of the action of learning and the generation of such, but in turn makes the model inflexible, less sophisticated, inherently shallow, static, and overall poorly designed on its own merit. Those solutions are temperamental, not permanent, nor it is in the long-run effective, though there is no doubt of its success in contemporary machine learning structures. As to push further, it is then to be believed that it is insufficient for such task. It is just similar as the replica problem of the imposed observations that we see. It feels like we are seeing patterns or uncovering something in data, or nature, usually can be agreed as such, but certain cases overvalued of such observations, while living in the world of which rules we create ourselves. 

The issues can be relieved, in certain sense, of certain solution, for example, to build the model within the newfound theory of modelling, of which the philosophy of constructing such `AI' model as to be from first component - i.e. ground zero, and of which the operating environment of said given model is rich enough as to interpret different types of rewards or similar-purpose responses. Or, in certain sense, couple uniform randomization, and replicate the genetic evolutional pattern, as seen in perhaps many researches on such algorithms of such, for example, \cite{stanley2002neat,stanley2004competitivecoevolution,stanley2002efficientRL,oneill2001grammatical,ryan1998grammatical,zhang2020retneat,khamesian2021hybridattentionneat,volna2005evolutionarynn,sher2013neuroevolutionerl}, and as to allow the observation in which the learning adaptation itself generates the `purpose', within reason and within probable causation chain (though, we must be careful of the overstretched implications). Or, perhaps there exists certain other structural construct that allows such, of philosophy and mechanism that embed such on its own. Structural and operational learning notion can be applied into many current frameworks of learning requirements, though most of the time we are entangled within fixed operational, dynamic structural (for example, SVM), or fixed structural, dynamic operational (any reasonable learning framework of existence with many-phase inference) learning. If we can resolve the natural occurrence, not necessarily as to not encode the notion of learning itself, since we are allowed up to certain point, to encode them to the model as of a shortcut and heuristic baseline, then the structural-operational learning sequence would be much obliged of help, and certainly would provide a larger and richer theoretical grounding toward the generality of the learning concept. 

\section{Conclusion}
In its final form, the paper serves mainly what it is intended of - to critique and analyse the non-trivial yet apparent nature of current existing framework in pursuit of the intelligence question, the lacking knowledge, the implicit and explicit holes in our understanding, and the misguided optimism, plus expectation and progression. Furthermore, it also lays out the foundation of thought, in which of the uncertainty currently presenting, partially formalize the philosophy and approaches that would be taken on, given the school of thought in Section~\ref{sec:3form}. If expanded, of make rigorous in which the substance is not eluded in complication, and the unpolished over-mathification of theoretical works, then potentially speaking, such framework introduced at the end would prove useful to the landscape in which of what we think of, as \textbf{machine}, and the modification of the \textbf{learning machine} on top. 

Nevertheless, we do not simply reject, less of our rhetoric in the paper itself, as to criticize current practices and development, the stagnation of theoretical studies, and so on. But rather, the focus is to reach toward those that claims of which overstretched the current ability at present in existing structures. Current systems are good of what it is --- more specifically, it is a \textit{revolution} in which absolved the debate between symbolic or neuronic system, into a more sophisticated and generalized form --- and thus advanced alongside computational development to be as it is. Nevertheless, we should still address the elephant in the room, as for AGI is unattainable because of the opaque nature of such terms, the philosophical gap between treatments and how we view such studies of creating intelligent, or in general, proxical-life\footnote{We say \textit{proxical}, derived from the word proxy, here is to mean of creating machine-encoded, system-embedded system of life-capable form. Such is to separate it from the natural evolution of life in its natural form, which is a very time-and-resource-intensive game to play in reality, hence, in a partial sense, an emulation of such complex occurrence.} subjects, the overreliance on the LLM mode of operation as potentially universal, in which fallacies from itself have already proven wrong, and the practical, mathematical and theoretical frameworks' inadequacy that is often not realized, but hidden in plain sight. Such is also why developments and further breakthroughs have stalled --- by the lacking of which heuristic cannot cover any more. It is then natural, such as in the paper that future work for us is to recognize such systematic mistakes, identifying the foundation's weakness, and either fix it, connect it, path it, or reframe entirely under different direction. Said directions realized in this paper might not make it into the future; perhaps being superseded by something much more sophisticated. That said, its end goal should still be conceptually interesting, and the critique would stay, much to the disdain of the deniers, and thus the purpose of this inquiry is completed. 
\section{Future works}

In future, following such conclusion, it is my purpose to extend those structures in which I laid out, verify their validity and sufficiency of such, and either replace them by something else, or proceed to construct them further on. That includes the theoretical system in which we can define, though temporarily, the research program in which start from ground zero of the constructs (i.e. the \textit{autopoietic system construction}), the definition of intelligence revisited and to reinterpret benchmarks and verification methods as of current, model definition and more works on the evolutional, percolation patterns of models itself, clarifying and putting in it the technical sufficiency for the neural network formalism, and many more works to come. 

\subsection{Limitations, Errors, and Author Contact}

Because it is inconclusive, mostly theoretical, and the author recognize inconsistency in certain parts of the paper, there are limitations and errors that would emerge in detail future revisions. Furthermore, because the paper also introduces conceptual frameworks for understanding artificial intelligence, and so on for future theoretical understanding, there exist errors and caveats of which should be addressed, or fixed if suitable reasons are to be provided. These are intended to guide readers in interpreting the materials, and to clarify the assumptions underlying the argument. 
\begin{enumerate}
    \item Certain conjectures rely on vague conceptual interception, while some hints at idealized assumptions. For example, in conjecture~\ref{conj:conjecture_AI_1} and conjecture~\ref{conj:conjecture_AI_2}, these are intentionally vague as to clarify in an operational sense, the perception of AI in modern standards. Those can be confusing, and there might exist fallacy within the conjecture itself. The Universal Approximation Theorem (UAT) and its simplified notion assume smooth, continuous functions over compact subsets, which may not hold for all real-world representation of data-encoding. 
    \item Many concepts are theoretical, and there are no sufficient guarantees in which they would be accepted or not superseded by better replacements. This accounts for the \textbf{dual formalism} listed in section~\ref{sec:3form}, and so on of the learning theoretic. In the appendix, the same goes for the layered Chinese Room formalism and its associated recursive threshold $Q$. 
    \item Architectural limitations, specifically the examples on GNN and others, rely on current theoretical and practical results, and may not generalize to future architectures, alternative encoding schemes, or any relevant architectures outside the author's knowledge. 
    \item The definition of artificial and intelligence are \textbf{inherently subjective}. While this paper attempts to formalize even the notion of subjectiveness in such definition, interpretations might differ across contexts, disciplines, observers, and so on. 
    \item Analogies such as "man-in-the-box", the argument about Searle, should be taken as \textit{heuristic pictures}, rather than literal cognitive model. This also hints at a potential structuralization of such concept in a detailed form. 
    \item Discussion on AGI and emergent intelligence also reflect \textit{current understanding, trends, hypes, directions}, and also philosophical reasoning, which may evolve overtime as new theoretical or empirical findings emerge. 
    \item Some claims, particularly regarding scaling laws, bottleneck phenomena, or functional expressivity, are derived from simplified or idealized settings, and thus is debated in the paper. For the neural scaling law, we directly argued it as a weak form of correlation via morphed reference space, though it does not inherently rule out the sufficiency in which such observation might help in practical settings. Thereby, it is also to be noted of those claims as to have controversial positions. 
    \item The results in certain cases would be regarded mainly as conceptual guidance rather than prescriptive implementations, as such, production of certain analyses, and verification, very much relies on practical and experimental designs, of which the paper does not cover. 
\end{enumerate}

All interpretations, derivations, and conjectures henceforth in this paper are the responsibility of the author. While cares have been attributed to provide as rigorous as possible reasoning and many fail-safes to be added, errors in logic, assumptions, or presentations (spelling or grammatical errors) may exist. It is then encouraged to critically assess, challenge, or refine the presented ideas, and also contribute correcting inconsistencies or presentation issues. For questions, discussion, suggestions, and fixes, the author can be reached at \href{mailto:fujimiyaamane@outlook.com}{this e-mail}. 

{
\footnotesize
\bibliography{references}
\bibliographystyle{iclr2025_conference}
}

\appendix
\section{Appendix. A potential answer to CRA}

Somewhat, in a way or another, a solution can be reached at least in terms of interpreting intelligence in a varied different way. Of such, there exists my blueprint notes of before, which outlined particularly interesting aspects and somewhat naive representation and ideas on a functional construct similar of those established above. While it is not perfect, and the line of logic leaves much to be desired, the idea is nevertheless worth of consideration. 

\subsection{The merits of the processing unit}

In fundamental neuroscience, one of the main question, that was answered figuratively and conceptually, is the question about the \textbf{merit of existence} of a separated organ that eventually, will exhibit controlling behaviours and might up to the point of intelligent behaviours. That is, it raises that 
\begin{quote}
    Protists and the simplest multicellular animals (sponges) display ingestive, defensive, reproductive, and other behaviours without any nervous system whatsoever, raising the question: what is the adaptive \textbf{value} of adding a nervous system to an organism?
\end{quote}

This point is, as it might sound, dilligently enough is very strong. We study intelligence as the essential part to search and create generally, a learnable system, adaptable system, and rational system, but to \textit{which stage}, \textit{which layers} and which \textit{form} would this be classified into? Are our systems generally similar to the Proties and multicellular animals, or have advanced to a greater abstraction level? 

\subsection{The layer dilemma - or Chinese Room Argument}

We pertain to the \textbf{layer dilemma}, or as generally recalled, the \textbf{Chinese room argument} [Searle, 1980]. If we choose to simplify partially our system to an IO conduct, then this argument is the first to be addressed. However, this will be done in a very problematic way. 

\begin{assumption}
Of the layer dilemma, for the set $\{L_i\}_{i\leq n}$ of the bottom-up categorization and measure up to $n$ layers, then $L_{n-1}$ do not "understand" what happens in $L_n$, but $L_n$ is of itself, with sufficient realization and connection. This means the hierarchical information pathway is one-way, bottom up. \footnote{Note that, in here, the presumptuous assumption is that each individual layers, or, the \textit{man in the box}, do not understand the higher scope he is received. Yet, he can understand what he has to do, hence the operation is still valid. However, understanding what was required and interpreted by the upper layer, simply makes it so that the man himself cannot interpret what was relayed, and hence do not possess any capable realization on such fact.}
\end{assumption}

However, if is so, then there exists the fact that we can define a condition $Q$ on such that, for a measurable complexity and `model capability' scale $A[p]$, then if $A[p]\geq Q$, the model escaped the potential restriction, and become recursive in nature. In principle, this such that the \textit{man in the box} recursively think outside what is conceived to be in his control. For every downstream, there exists an encoding to the lower dimension. Such downstream inevitably will result in an exposure to the data, in some form or another, that can be decoded. If, the model capability for the outermost layer, $L_n$, satisfies the inequality, then it can, indeed, escape, and interfere directly to a certain extent of the actual `world' layer $L_0$ it receives downstream to, which is, where we feed it the data and else, isolated from the model. This is, in principle again, totally a hypothesis, but then, we need to have a boundary to restrict this. 

\begin{conjecture}
    Suppose there exists a measure $\mathcal{M}[L_i \in A_n]\in \mathbb{R}^{p}$, usually for $p=1$ as scalar for any model configuration $A_n$. Then, if $\mathcal{M}[L_i]\geq Q$ for $Q\to \mathrm{dim}(p)$, then there exists a recursive pattern $\mathrm{Re}(A)$ such that it enables the model's ability to 'learn' of the upper $L_{i-1}$ layer.
\end{conjecture}

\begin{figure}
    \centering
    \includegraphics[width=0.85\textwidth]{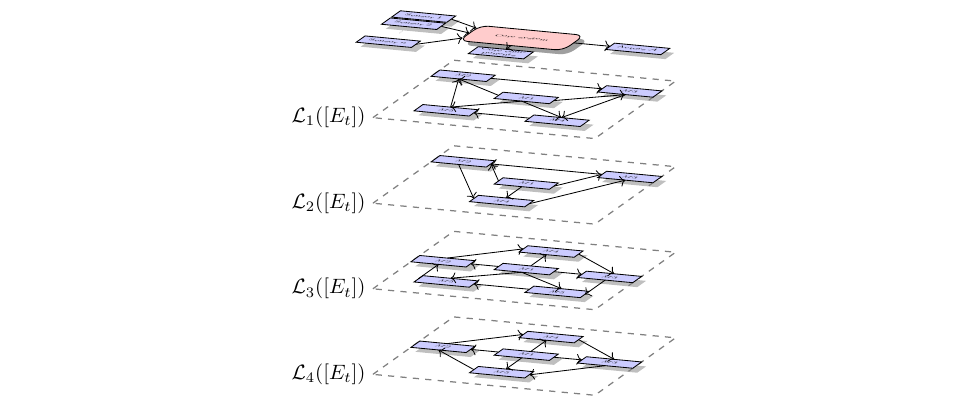}
    \caption{A loose illustration of the layering principle. The lower layers do not know what is on the upper part. However, they can receive potential downstream input features, and hence work accordingly. This is the basis of \textbf{abstraction by layers}, thus stating that \textit{the man in the box do not know, and do not understand, anything but his circumstances and his current capability}. Nevertheless, it still fulfils its role in the lower level hence forth, and its operation only in such range of the layer.}
\end{figure}

Hence, the answer in such form is reached, such so to address the fact that the operating machine receives downscale information within its own embedding, hence is typically restricted to that embedding layer alone. Furthermore, it also usually satisfies, since we categorize and generalise the layering, as according to the generalisability and abstraction value of each. Therefore, conspicuously, most of the time, the layer will hold, but under configuration, then it might jumps off the layer without any re-categorization method.  

The layering theory also is an answer to the Chinese room argument, in the sense that there exists levels of which the abiding room is taking into account. Hence, for a person living in a scenario of input-output interaction, it is well-conceived that the `I', within such capability, within such restriction, and being categorized as 'lower', do not possess the ability to look outside the room itself. A possible hypothesis, would be in analytical form, there exists a separated evaluating space, where such layering principle is demonstrated by dimensional increases, and their functional downstream compression function. 

\begin{hypothesis}
    Given the set $\{L_i\}_{i\leq n}$, there exists a dimensional analytic encoding $\mathsf{Comp}^{m}$, such that there also exists certain functional downstream compression function $F_{DC}$ that maps $\{L_i\}\times \mathsf{Comp}^{m}\to\{L_i\}\times \mathsf{Comp}^{m-1}$, essentially rescaling the dimensionality. 
\end{hypothesis}

Of course, this is all conceptual. We still need to define the measure $\mathcal{M}$, the supposed inequality, the supposed construction, and furthermore, the complete test implementation that can be perhaps focused on and verify the principle. Nevertheless, this might prove useful, or at least in a direction of analysis that will either falsify entirely this approach, or partially, thus makes use of this to formalize new framework on understanding such problem\footnote{Especially this, for the variable $Q$, which then by default of this, must either be formulated correctly in speciality, or rather, tested empirically.}. Nonetheless, such notion makes sense, for Searle-in-the-box to only be `aware', of the arbitrary sense of the word `aware' attributed of the analogous human awareness of what of his own domain, but not of any external or higher concept. It is apparent of human, too, as we can only semi-imagine the higher-level being, in philosophical inquiry, but nothing else --- however when it comes to lower subjects, then it is surprisingly rich of certain interpretations. Such inquiry will remain valid, as for all of such layering scheme we can create. Such answer leads to two development. We can identify the layer of human, and try to build up layer-by-layer up there. Or, we might want to believe, there exists this layer, $L_{\epsilon}$, of which such restriction of top-down interpretability becomes redundant, in which the subject leaps out of its layer to the above. What of the two is more probable? We do not know.  

In one way or another, CRA infers with the option of an \textit{input-output} procedure. In such case, CRA clearly tells us that a simple, mundane notion of an input-output machine which 'does its tasks' would not be sufficient of receiving the clarification and qualification for being intelligent. Which, by all means and purposes, are true. The construct if given in such circumstances, of the thought experiments, represents, if we took out the part where the argument said Searl is supposed to be everything a computer can be, then it's true that such constructs are false in its claim that it can 'understand'. In fact, relatively simple, a given input-output mechanism, from what was observed from the outside, do not exhibit anything, and do not have the ability to even \textit{think}, regardless that is valid of such. If we are to stand by Descartes arguments, then it is even more of the truth - the system in which the thought experiment was conducted provides it with no capabilities of any such.

Then what would be of the Chinese Room Argument that is worth it to dissect? Well, firstly, the claims of, at least in the acute interpretation - the man in the box is supposed to be everything a computer can be is \textit{false}. IF we are to stand by our construction of facilities, then we are innately arguing about such facilities, and not the processes, and the underlying operations itself. Rather, we are complaining that the machine is not capable enough, which is true. But we also, to a given reference point, pointing to the \textbf{existential facilities} instead of the arbitrary, yet reasonable operational facilities instead for the comparison. And in fact, if we think of it in the layer construction, it makes more sense - each layer is classified given its arbitrary for now, an interpretation thereof. Such interpretation is contained for such layer, and hence cannot be thoroughly or at least in a glance, interpreted by lower-layer components. This construct offers a one-way restriction on the property of interpretability. In such case, the \textbf{System Reply} is partially right - the man inside can not be, by all means and purposes, understand what does it mean by even 'English', or 'Chinese'. And even if such English 'understanding' is embedded in the reasonable interpretable space of Searl in the room, then Chinese would appear to be entirely unknown, \textit{unless} there exists a helper tool to resolve the situation of undefined operation. Indirectly, this prevents a Descartes situation from happens - an undefined situation with particular way to resolve.

But more than that, what constitute the notion of understanding? Taken only from the setting of the thought experiment, we cannot do but deceptively assume that understanding a language is similar to giving it certain ruleset to transfer from this word to others word only. By that, converting from base-2 to base-10 works the same - you know how to do it, yet there are things that constitute the philosophical, higher-level notion of 'understanding' in such that allows you to actually understand the conversion - otherwise, the conversion is a blind matching. In fact, given the setting, would a conversion rulebook exists for such language? Language is, by itself, a very high-level concept. Translating from text to texts requires it not only to provide the definition matching, which could be reasonably identified by such notion like the rulebook mention, but the interpretation of the string is dictated by the logic of the language - the context in which it appears, the logical conformation that it contributes, the grammatical structure that makes sense of what is said and what is transferred, and else. Language itself, is a medium of information exchange, by one of its definition. A conversion does not constitute an understanding. And if the argument going back and forth is that Searl can somehow figure it out the patterns mean something, then it violates what I called the \textit{principle of externality} - the 'lower components' up to a given point in which the \textit{law of recursive immergence} does (not) apply, cannot implement its higher constructions. Then, the Chinese Room Argument can be interpreted, in somewhat meagre form, the argument against telling the current conformation as able to understand, while it is not.

\subsection{The value of intelligence}

\section{Appendix. Argument against the Gödelian argument}

In 1961, J. R. Lucas presents the Gödelian argument against the existence of a "strong" AI. His proof is based on Gödel theorem, which is stated as followed:

\blockquote[Lucas, 1961]{In any consistent system which is strong enough to produce simple arithmetic there are formulae which cannot be proved-in-the-system, but which we can see to be true. Essentially, we consider the formula which says, in effect, "This formula is unprovable-in-the-system". If this formula were provable-in-the-system, we should have a contradiction: for if it were provable-in-the-system, then it would not be unprovable-in-the-system, so that "This formula is unprovable-in-the-system" would be false: equally, if it were provable-in-the-system, then it would not be false, but would be true, since in any consistent system nothing false can be proved-in-the-system, but only truths.}
This theorem holds for all formal systems which are consistent, adequate for simple arithmetic, and shows that those formal systems are incomplete, with some fact being true, but unprovable.
\blockquote[Lucas, 1961]{It is of the essence of being a machine, that it should be a concrete instantiation of a formal system. It follows that given any machine which is consistent and capable of doing simple arithmetic, there is a formula which is incapable of producing as being true\dots}

Further argued, he then comes to such conclusion that no machine can be a complete or adequate model of the mind, since "the mind are essentially different from machines". Lucas's defenders, Roger Penrose, also state in his \textit{Shadow of the Mind} (1994). A human mathematician, if presented with a sound formal system $F$, could argue as followed:
\blockquote[Penrose, 1992, 3.2]{Though I don't know that I necessarily am $F$, I conclude that if I were, then the system $F$ would have to be sound and, more to the point, $F'$ would have to be sound, where $F'$ is $F$ supplemented by the further assertion "I am $F$"\footnote{The phrase "I am $F$" is merely a shorthand for "$F$ encapsulates all the humanly accessible methods of mathematical proof"}. I perceive that it follows from the assumption that I am $F$ that the Gödel argument $G(F')$ would have to be true and, furthermore, that it would not be a consequence of $F'$. But I have just perceived that "If I happen to be $F$, then $G(F')$ would have to be true", and perceptions of this nature would be precisely what $F'$ is supposed to achieve. Since I am therefore capable of perceiving something beyond the powers of $F'$, I deduce that I cannot be $F$ after all. Moreover, this applies to any other system, in place of $F$.}

By default, the argument supplemented from Penrose raised the contradiction of proof-ness. The Gödelian argument implicitly creates layers, and levels, on which one puts those languages they are abided to seem fit of their expressions on the shelf, by the order of \textit{effectiveness}. Such notion then, would make the advancement of machine to human seems perpetually, unsophisticatedly, inoperable and impossible in essence. Lucas argument, just as Searle, also claim that it is all the computer can do, of which the system itself is inherently useless of hosting such entity. However, artificial intelligence, as for now, using this term since chapter 3 which is not yet here, is not a computer in its form. We say, however, for an \textit{artificial intelligent subject with computers as its existential facilities}, not the computer itself. This open up the fact that the notion of computer we are having right now, are also limited to the kind of classical computer, and not taking into account of any such similar `computing architecture' of framework that might differ from such understanding. If so, then CRA is only partially right. But partially wrong since the comparison is limited to a form of internal structure in a well-formed system. That is to said - we need to create the (a) construct(s) that exceed(s) such argument. The problem is, how?
\subsection{Remark}

Before even taking a stance on such argument, what is the meaning and interpretations, as well as ostensively why it is even important to divulge into such point? The answers might be a bit difficult.

Human is variedly different from machine, for the current time with all the knowledge at present. Truth to take, the action of writing this itself is part of the endeavour to discover one's self, or rather, to understand $F$ with the assertion of 'I am $F$', for now, that we can, and is doing. By the language and construction of contemporary and propositional logic, a machine cannot do that.\footnote{In general, we cannot even say that it is true of the truth that human actually differs from what is proposed to be perceiving $F'$ being $F$. For human understanding of ourselves alone, we are trying to fit it into the interpretation and the rough 'understanding' of human itself. That is, there exists the space of reason and argument of a scientist, which interpretation follows. If, supposedly, this interpretation is strong enough, then we might be able to perceive and understand ourselves from ourselves - a looped interpretability. This mechanism, if ever, is not well understood if exists.}

However, if the converse situation happens, where we cannot totally perceive what we actually think, and how it is formed - per metric, being either consciousness, or one's self, surprisingly, it does not support the previous argument from an intuitive view (Bear in mind that this is a non-rigorous study). If stays rigid as it is, not counting being dynamic as we want, the model created from a human being can only imitate and represents what directly is entailed in the human mind of interpretation and logics. But logic and interpretation is a construct of the mind, for all intents and purposes, to directly infers to the physical world, the living world. However, if one is to use such inference on itself, for example, examining the brain itself, then to a certain point, what can be deduced from such observation can only fit in the interpretation space of what its creator, the human brain itself, can contrive. Thereby, we might conclude that figuratively, even human cannot understand human itself, from certain perspective. But the quality of succinctly interesting loop is to be taken seriously. The point now is, what type of construct, even logic, would be sufficient of taking the understanding, and will it make uses of the looped behaviours? By that, we then argue superficially that anything that relies on the machine cannot model the human existence and conscience itself. There are some assumptions thereof in the argument:
\begin{itemize}[noitemsep,topsep=1pt]
    \item Existences and the state of the world are in fact, modelled in mathematics, for one way or another. This is to facilitate the use of formal system in the argument. Everything is a set of rules, in which things operates.
    \item A machine per its definition, cybernetic machines are of all expressed by the single principle that it is born out of a formal system itself.
    \item Truth is the finite quantity that exists in such formal system, and is absolute. 
    \item The mind is an entity of which is inherently different from the logic of formal system.
\end{itemize}
Those fundamental, overlapping assumptions make up the bulk of the Gödelian argument, from the surface. However, is that true of all the merit? \footnote{It turns out, however, the Gödelian argument has various proponents and opponents, and there are arguments of it being false. See \cite{cogprints553} for such argument, but it can be simplified as this. The Gödelian argument makes use of two assumptions: $G(F')$ is true for a Gödelian statement, and $F'\not\vdash G(F')$ for $F'\not \vdash G(F')$ for $F'$ being "I am $F$" with added semantic. Then, the statement on $G(F')$ is true is nothing but a \textit{satisfaction} claim, of meta-mathematical assertion which can be reduced to $\mathcal{I}\vDash G(F')$ is true for given interpretation $\mathcal{I}$. Thereby, there exists no contradiction thereof.}

\end{document}